\pdfoutput=1
\documentclass[11pt]{article}

\usepackage[margin=1in]{geometry}
\usepackage[T1]{fontenc}
\usepackage{lmodern}
\usepackage{microtype}
\usepackage{float}

\usepackage[english]{babel}
\usepackage{amsmath,amssymb}
\usepackage[version=3]{mhchem}

\usepackage{graphicx}
\usepackage{float}
\usepackage{placeins}
\usepackage[format=plain,justification=justified,singlelinecheck=false,
            font={stretch=1.125,small,sf},labelfont=bf,labelsep=space]{caption}

\usepackage[hidelinks]{hyperref}
\usepackage[numbers,sort&compress]{natbib}

\usepackage{authblk} 

\title{Optimal message passing for molecular prediction is simple, attentive and spatial}

\author[1]{Alma C.\ Castañeda-Leautaud\thanks{Corresponding author: \href{mailto:acastanedaleautaud@ucsd.edu}{acastanedaleautaud@ucsd.edu}}}
\author[2]{Rommie E.\ Amaro}

\affil[1]{Department of Chemistry and Biochemistry, University of California, San Diego, 4238 Urey Hall, La Jolla, CA 92093}
\affil[2]{Department of Molecular Biology, University of California, San Diego, 4238 Urey Hall, La Jolla, CA 92093-0340}

\date{\today}  

\begin{document}
\maketitle

\begin{abstract}
Strategies to improve the predicting performance of Message-Passing Neural-Networks for molecular property predictions can be achieved by simplifying how the message is passed and by using descriptors that capture multiple aspects of molecular graphs. In this work, we designed model architectures that achieved state-of-the-art performance, surpassing more complex models such as those pre-trained on external databases. We assessed dataset diversity to complement our performance results, finding that structural diversity influences the need for additional components in our MPNNs and feature sets. 

In most datasets, our best architecture employs bidirectional message-passing with an attention mechanism, applied to a minimalist message formulation that excludes self-perception, highlighting that relatively simpler models, compared to classical MPNNs, yield higher class separability. In contrast, we found that convolution normalization factors do not benefit the predictive power in all the datasets tested. This was corroborated in both global and node-level outputs. Additionally, we analyzed the influence of both adding spatial features and working with 3D graphs, finding that 2D molecular graphs are sufficient when complemented with appropriately chosen 3D descriptors. This approach not only preserves predictive performance but also reduces computational cost by over 50\%, making it particularly advantageous for high-throughput screening campaigns.
\end{abstract}

\section{Introduction}

A neuron in deep learning (DL) is simply a linear equation, but when multiple neurons join together with an activation function (e.g., Sigmoid or ReLu) they form a non-linear transformation. Supported by the universal approximation theorem, neural networks (NNs) can virtually adapt to any kind of data\cite{maiorov_lower_1999}. 

Graph Neural Networks (GNNs) rely on neural networks to operate on graph-structured data \cite{velickovic_everything_2023}. Since molecules are intrinsically graphs, where atoms are nodes and bonds are edges, GNNs are capable of learning any molecular structure and transform it into separable patterns, facilitating tasks such as toxicity prediction, activity classification or atom-level predictions\cite{kim_molnet_2022}. GNNs have become widely applied in medicinal chemistry since the re-purposing of Halicin, a newly discovered antibiotic using a directed Message-Passing Neural Network (MPNN)\cite{stokes_deep_2020}.

MPNNs are a type of GNN, where the “message” concept refers to the metadata abstracted from each atom (and sometimes edges) and passed iteratively to adjacent nodes using a mathematical operation that combines each node embedding into an output of defined size. This operation is known as the aggregation function \cite{gilmer_neural_2017}. This function imposes a direction possibly harming the molecular representation from a chemical point of view. Molecules lack a defined start and end, making the imposition of an A-to-B connection counterintuitive. Additionally, covalent bonds are fundamentally symmetric, representing mutual interactions rather than a one-way flow of information. One strategy to circumvent directionality in MPNNs is to aggregate the message in both directions for each node, effectively obtaining a bidirectional message passing.

MPNNs are usually classified into three main flavors\cite{bronstein_geometric_2021}: Message-Passing where nodes generate messages based on their own features and those of their neighbors, aggregate incoming messages, and update their representations through learnable functions allowing learning relational inductive biases \cite{gilmer_neural_2017, battaglia_relational_2018}. The second includes Graph Attention Networks (GATs), which incorporate attention weights during data transfer between nodes\cite{velickovic_graph_2018, brody_how_2022}. The last flavor is constituted by Graph Convolutional Networks (GCNs) which exploit convolution normalization to add a penalty on highly connected atoms\cite{kipf_semi-supervised_2017}.

We noticed that the combination of the three flavors into hybrid architectures is mathematically feasible and could exploit the advantages that each method has to offer. Additionally, Message-Passing can be simplified by avoiding the insertion of raw self-nodes after message processing. We hypothesize that eliminating redundant feature amplification is unnecessary for small graphs, such as molecules, which typically have 20-70 heavy atoms\cite{isert_qmugs_2022}.

Traditionally, GNNs represent molecules using 2D features including atomic number, hybridization, bonds conjugation states, bond types, number of hydrogen bond donors and acceptors, fraction Csp3 and LogP \cite{boulougouri_molecular_2024,chen_learning_2021,jiang_graph_2020, liu_chemi-net_2019,yang_analyzing_2019, rittig_graph_2023}. The use of these types of features is probably due to the ease of extraction provided by RDKit modules, which we have noticed do not include chemically meaningful element-like descriptors such as van der Waals radius, electronegativity, and dipole polarizability \cite{landrum_rdkitrdkit_2025}. While these approaches are capable of encoding atom connectivity, they fail to capture stereochemical properties critical for drug discovery such as steric hindrance and radius of gyration. Stereochemistry is crucial when designing new drugs, as even a single chiral center can significantly influence potency toward a target of interest \cite{mcconathy_stereochemistry_2003}. 

In this study, we: 1) constructed MPNNs with global and node-level outputs to analyze their performance, 2) offer a re-interpretation of message-passing in MPNNs by means of modifying how the message is passed and assessing node self-perception, attention and convolution mechanisms and 3) evaluated novel 3D features, widely used 2D features and added element-like 2D features on three different datasets applied to benchmark performance on drug-discovery campaigns.

Finally, we make available a code that readily reads and transforms molecules in SMILES (Simplified Molecular Input Line Entry System) format to 3D graphs with an option to visualize predictions at the node level using colormaps, helping to interpret the patterns recognized by the deep learning model, effectively facilitating machine generating insights for chemists.

\section{Theoretical Foundations of Graph Neural Networks as Molecular Representations}

\subsection{Directed and Undirected Graphs} 

We can mathematically represent a directed graph in terms of the number of edges (\( i = 1, \dots, N_e \)), where information flows from source (\( s_i \)) to destination nodes (\( d_i \)) and a collection of global (\( g \)) and edge (\( e_i \)) features are included:

\[
G = \left( g, e_i, s_i, d_i \right)_{i=1,\dots,N_e}  \tag{1}
\]

The connectivity map is stored in an adjacency matrix 
\(
A \in \mathrm{R}^{n \times n}
\), which, in the case of a molecule, is represented by a square matrix with a length equal to the number of constituent atoms.

One way to artificially create an undirected graph in MPNNs is to create a symmetric adjacency matrix. Here, for every edge (\(s_i, d_i)\) there would be a corresponding reverse edge (\(d_i, s_i)\). Alternatively, a two-way direction (bidirectional) can ensure that the source would also perceive the destination nodes \( s_j \to d_j \) and \( d_j \to s_j \).
An overlooked aspect of the message concept in MPNNs is the notion of the self-node. This is particularly useful for handling vague node representation caused by large graphs where information vanishes as it propagates. To include the self-nodes in the adjacency matrix A, we can modify it by adding an identity matrix I to A. Alternatively, node features can be concatenated into the aggregated messages to ensure self-perception is retained.

\subsection{Properties of the GNNs} 
The architecture of a GNN must satisfy certain conditions to preserve its structural integrity during data processing: 
\subsubsection{Permutation Invariance}

Indicates that the result after a permutation operation remains unchanged. This means that the function output is independent of the order of the data input. Examples of such functions include summation, averaging, and maximization, which are termed "aggregation functions" \( x_{NX} \) \cite{bronstein_geometric_2021}. In terms of the adjacency matrix \( A \), the permutation matrix \( \rho \), acting on the node features vector \( X \), is permutation invariant if:

\[
f(\rho X, \rho A \rho^T) = f(X, A) \tag{2}
\]

where the term \( \rho A \rho^T \) represents the aggregation operation acting on the matrix \( A \).

\subsubsection{Permutation Equivariance}

This property ensures that any permutation applied to the input is reflected in the output in the same manner. Permutation equivariant functions can be represented as \( \phi \) \cite{bronstein_geometric_2021}. In terms of \( A \), we can express the equivariance condition as:

\[
f(\phi X, \phi A \phi^T) = \phi f(X,A) \tag{3}
\]

This property has the advantage of incorporating a sense of locality, ensuring that nearby processed nodes exhibit similar transformations, i.e., \( \phi(x_i) \approx \phi(x_j) \). Convolutional Neural Networks (CNNs) serve as a prominent example of permutation equivariant operators \cite{fukushima_neocognitron_1980}.

\subsubsection{Relational Inductive Bias}

Means that locality is reinforced by imposing constraints on interacting nodes during the learning process \cite{battaglia_relational_2018}. Under this context, information is biased toward learning from connected nodes, a key characteristic of Message Passing Neural Networks (MPNNs), though not exclusively so. Molecules particularly benefit from this property, as it enables the detection of functional groups associated with specific chemical properties, such as solubility or toxicity.

\subsection{Defining GNNs}

Having listed the characteristics that GNNs must obey, allows us to introduce a mathematical expression to formalize them. A minimal expression for a GNN algorithm that operates on nodes only, \( f(x) \), is:

\[
f(x)= \phi \left( x_i , \sum_{j \in N_i} x_j \right)  \tag{4}
\]

where the target node ($x_j$) is aggregated to node $x_i$ and $\phi(x)$ could be a multilayer-perceptron (MLP). 
Additionally, global $u$ and edge $e_{ij}$ features can be incorporated. 
An example could be:

\[
f(X)= \phi \left( x_i , \sum_{j \in N_i} (x_j, e_{ij}, u) \right) \tag{6}
\]

GNN architectures can be categorized into three main types \cite{bronstein_geometric_2021}:

\subsubsection{Convolutional GNNs.}

Convolutional GNNs, such as Graph Convolutional Networks (GCNs) \cite{kipf_semi-supervised_2017}, employ a convolution-like aggregation function across node neighborhoods. Before aggregation, the features of the connected nodes are normalized using the factor \( C = \frac{1}{d_i d_j} \), where $d_i$ and $d_j$ denote the degrees of nodes i and j, respectively. This escalation balances nodes contributions.

Thus, the aggregation function in a convolutional GNN can be expressed as:

\[    f(X) = \phi \left( x_i , \sum_{j \in N_i} C(x_j) \right)
    \tag{7}
\]

GCNs have demonstrated effectiveness in processing molecular structures using this simple normalization factor, making them the fastest among the GNN categories \cite{deng_xgraphboost_2021, velickovic_everything_2023}.

\subsubsection{Attentional GNNs.}

Attention mechanisms were originally designed for sequence-to-sequence models, which allowed different importance values to be assigned to elements within a sequence, improving sensitivity to contextual information \cite{bahdanau_neural_2016}. This concept was later adapted to GNNs as Graph Attention Networks (GATs) \cite{velickovic_graph_2018}.

In the classical GAT model, the attention coefficient \( a(x_i, x_j) \) between adjacent nodes is computed as:

\[
a(x_i, x_j) = \frac{\exp(\text{LeakyReLU}(a \cdot [\phi x_i \| \phi x_j]))}{\sum_{j' \in N_i} \exp(\text{LeakyReLU}(a \cdot [\phi x_i \| \phi x_{j'}]))}  \tag{8}
\]

where \( a \) is a learnable attention vector applied to the concatenation of source and target node representations, each of which has been independently processed. The associated weights and biases are treated as trainable parameters. A LeakyReLU is an activation function employed to amplify neurons with promising parameters while assigning small gradients for negative ones, thus avoiding elimination. A SoftMax normalization is applied following activation\cite{bridle_training_1989}. Including  \( a \) in the GNN expression yields, 

\[
f(X) = \phi \left( x_i , \sum_{j \in N_i} a(x_i, x_j) x_j \right)  \tag{9}
\]

Additionally, multiple attention heads can be incorporated into a GAT by computing \( K \) different attention weights and concatenating them to create the final representation:

\[
f(X) = \phi \left( x_i , \bigg\|_{k=1}^{K} \sum_{j \in N_i} a_k(x_i, x_j) x_j \right)  \tag{10}
\]

\subsubsection{Message-Passing}

MPNNs operate in a two-step process: message computation followed by the passing \cite{gilmer_neural_2017}. In the message computation step, feature embeddings from pairwise neighbors are processed with a neural network. These embeddings are then aggregated to each source node, processed and then aggregated to the destination nodes, completing one passing. After sufficient iterations, every node encodes information from all other nodes in the graph, making MPNNs particularly effective for tasks requiring long-range dependencies, such as molecular property prediction (e.g., solubility estimation).

Mathematically, an MPNN is expressed as:

\[
f(X) = \phi \left( x_i , \sum_{j \in N_i} (x_i, x_j) \right)  \tag{11}
\]

where \( (x_i, x_j) \) represents the processed message aggregated to node \( i \) along with its self-node features \( x_i \). At each iteration, messages are recalculated and propagated, making this approach computationally slower compared to other GNN categories. However, this iterative process enables the learning of more detailed node embeddings.

After defining the categories of GNNs we can observe that using one does not preclude the integration of others. Specifically, convolutional normalization factors and attention weights can be extracted and incorporated into the message-passing framework. The primary objective of this work is to evaluate how MPNNs, enhanced with convolution and attention mechanisms, influence binary prediction tasks in the context of molecular property prediction.

\section{Experimental Design and Methodology}
\subsection{General Model Architecture}
We constructed five variations of MPNNs that included atom, bond and molecular features to explore the effects of attention, convolution and self-node detachment, using undirected and bidirected graphs. The architectures followed the MetaLayer style \cite{battaglia_relational_2018}, where our main focus was in the node block, leaving the rest of the code intact (Message and global processing) (Fig. 1).

\begin{figure}[h]
\centering
  \includegraphics[height=8.3cm]{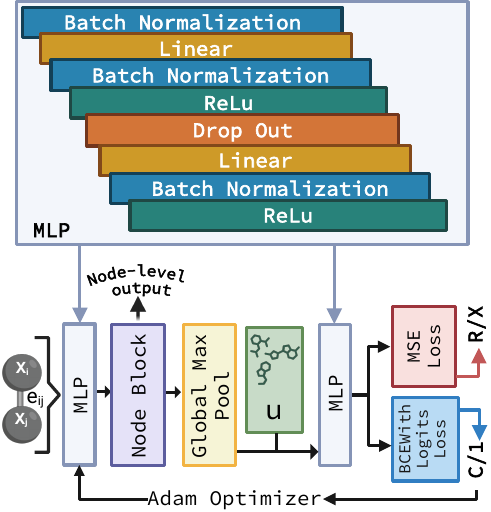}
  \caption{Diagram illustrating the general architecture of our MPNN for molecular classification. The message is processed in a batch-wise manner including source ($x_i$), destination ($x_j$) nodes, and edge attributes ($e_{ij}$), which are first processed by a Multi-Layer Perceptron (MLP), detailed in the top inset. The message-passing module depends on the tested model and optionally outputs node-level embeddings for colormap visualization. A global max pooling operation aggregates the node-level outputs into a single molecular representation, which is then concatenated with global features. This pooled representation is subsequently passed through another MLP to produce a single scalar logit in the case of classification task (C/1), which is processed by the Binary Cross-Entropy with Logits Loss (BCEWithLogitsLoss) operator or a scalar value for regression (R/X) with error calculated using Mean-Square Error Loss (MSELoss). The gradient is optimized using Adam.}
  \label{fgr:example}
\end{figure}

The core algorithm around the node block follows the general form:

\begin{equation}
    f(X) = MLP \left(u, \max_{u \in N_u} \phi_u[MLP(x_i, e_{ij}, x_j)] \right)
    \tag{12}
\end{equation}

Where an MLP processes the concatenated global features with the processed message using scatter max-pooling. The modifications that define our models are done at the node block ($\phi_u[\cdot]$), which integrates the passing procedure of the message. In all of our models the message encompasses nodes and edges, $(x_i, e_{ij}, x_j)$.
 
\subsubsection{Node block: Bidirectional MP Model (BMP)}

In this model we use a directed adjacency matrix and integrate a double-directed aggregation function to make sure each node perceives another. In a classical MPNN, raw self-node features are integrated after message passing. In this case however, we avoid mixing raw with processed embeddings and aggregate the message to the according atom indexes using a scatter maximization pooling, effectively taking the maximum incoming value to each considered node. Both backward and forward embeddings are concatenated and passed through an MLP,
\begin{equation}
    \phi(x_i, e_{ij}, x_j) = \mathrm{MLP}\left( 
        \max_{j \in N_i} \mathrm{MLP}(x_i, e_{ij}, x_j), \;
        \max_{i \in N_j} \mathrm{MLP}(x_i, e_{ij}, x_j)
    \right)
    \tag{13}
\end{equation}

The key operation that defines bidirectionality is that $\max_{j \in N_i}$ aggregates messages from neighbors of node $i$ and $\max_{i \in N_j}$ aggregates messages from neighbors of node $j$.

\subsubsection{Node block: Bidirectional-MP with Self-Nodes (BMP + SN)}
The BMP+SN is similar to the BMP and is characterized for explicitly incorporating raw self-node features after the message has been passed in both directions,
\begin{equation}
    \phi[x, x_i, e_{ij}, x_j] = \mathrm{MLP}\left(
        x, \max_{j \in N_i} \mathrm{MLP} (x_i, e_{ij}, x_j), \;
        \max_{i \in N_j} \mathrm{MLP} (x_i, e_{ij}, x_j) \right)
    \tag{14}
\end{equation}

\subsubsection{Node block: Undirected-MP (UMP)}

This model uses a symmetrical adjacency matrix with duplicated connections to create undirected graphs. We use the node block from the \textit{MetaLayer} in \textit{pytorch.geometric} \cite{battaglia_relational_2018} that included raw self-node features. Different from the rest of our models this architecture uses global mean-pooling, instead of max-pooling.
\begin{equation}
    \phi[x, x_i, e_{ij}, x_j] = \text{MLP} \left( x, \; \frac{1}{|N_j|} \sum_{i \in N_j} \text{MLP}(x_i, e_{ij}, e_{ji}, x_j) \right)
    \tag{15}
\end{equation}

\subsubsection{Node block: Convolutional-Bidirectional-MP (CBMP)}

We sought to combine the convolutional flavor with message-passing. This model includes a convolution normalization (see equation 7) applied to the message in the BMP framework (equation 13),
\[
\phi(x_i, e_{ij}, x_j) = \textit{MLP} \left[ 
\max\limits_{j \in N_i} \frac{MLP(x_i, e_{ij}, x_j)}{d_i d_j}, 
\max\limits_{i \in N_j} \frac{MLP(x_i, e_{ij}, x_j)}{d_i d_j} 
\right] 
\tag{16}
\]

\subsubsection{Node block: Attentional-Bidirectional-MP (ABMP)}
The Attentional-BMP model applies an attention mechanism after message processing. Building on the GAT framework described in Equation~8, we extend the mechanism to incorporate edge features alongside source and target node representations. 
\[
a(x_i, x_j) = 
\frac{
    \exp\!\left(\mathrm{LeakyReLU}\!\left(a \cdot \left[ \phi x_i \, \Vert \, \phi e_{ij} \, \Vert \, \phi x_j \right]\right)\right)
}{
    \sum_{j' \in N_i} \exp\!\left(\mathrm{LeakyReLU}\!\left(a \cdot \left[ \phi x_i \, \Vert \, \phi e_{ij} \, \Vert \, \phi x_j \right]\right)\right)
}
\tag{17}
\]

Additionally, we replace the original summation step in GAT with max-pooling, enabling a message-aware attention mechanism that operates on processed messages before propagation.

\begin{equation}
    \phi(x_i, e_{ij}, x_j) = \textit{MLP} \left[
    \begin{array}{c}
        \max\limits_{j \in N_i} a(x_i, e_{ij}, x_j) \, MLP (x_i, e_{ij}, x_j) \\[8pt]
        \max\limits_{i \in N_j} a(x_i, e_{ij}, x_j) \, MLP (x_i, e_{ij}, x_j)
    \end{array}
    \right]
    \tag{18}
\end{equation}

\subsubsection{Node block: Attentional-Bidirectional-MP + Self Nodes (ABMP+SN)}

Finally, we sought to analyze the effect of combining the ABMP model with raw self-node features, investigating whether the increased performance correlates with the added complexity, in comparison to the single-component additions made to the BMP model.

\begin{equation}
    \phi(x, x_i, e_{ij}, x_j) = \textit{MLP} \left[
    \begin{array}{c}
        x \\[8pt]
        \max\limits_{j \in N_i} a(x_i, e_{ij}, x_j) \, MLP (x_i, e_{ij}, x_j) \\[8pt]
        \max\limits_{i \in N_j} a(x_i, e_{ij}, x_j) \, MLP (x_i, e_{ij}, x_j)
    \end{array}
    \right]
    \tag{19}
\end{equation}

\subsection{Datasets as Benchmkarks}
MoleculeNet\cite{wu_moleculenet_2018} and BindingDB (\cite{gilson_bindingdb_2016}) datasets were selected for benchmarking based on experimentally determined biological parameters relevant to drug discovery:

\begin{itemize}
    \item \textbf{BACE Dataset:} The BACE dataset in its qualitative (binary label) mode, composed of 1513 drugs tested for their activity against the $\beta$-secretase 1 receptor.
    \item \textbf{Blood-Brain Barrier Penetration (BBBP) Dataset:} A dataset containing binary labels of blood-brain barrier permeability for 2039 compounds.
    \item \textbf{TRPA1 Dataset:} A dataset retrieved from the BindingDB database , containing 3020 IC$_{50}$ values targeting the Transient Receptor Potential Ankyrin 1 (TRPA1) ion channel.
    \item \textbf{Lipophilicity Dataset:} A dataset comprised of 4200 molecules with experimentally tested octanol/water distribution coefficient(logD at pH 7.4) with values ranging -1.5 to 4.
\end{itemize}

\subsection{Computational Environment}

All experiments were conducted on a system equipped with a 13th Gen Intel Core i9-13900H CPU (14 cores, 20 threads, 5.4 GHz), an NVIDIA GeForce RTX 4070 Laptop GPU (8 GB VRAM, CUDA 12.7), and 32 GB of DDR4 RAM. 

The code is implemented in Python using PyTorch and PyTorch Geometric and is available in our GitHub repository \href{https://github.com/chemdesign-accl/BMPs}{https://github.com/chemdesign-accl/BMPs}. Most features were processed using RDKit \cite{landrum_rdkitrdkit_2025}, and Mendeleev \cite{mentel_mendeleev_2014}. Hyperparameter optimization was performed using the Optuna package \cite{akiba_optuna_2019}. We used NumPy, Pandas, MolVS, and Matplotlib for metrics, visualization, and data preprocessing \cite{harris_array_2020, hunter_matplotlib_2007, mckinney_data_2010}.

\subsection{Molecules as 3D Graphs}

As part of the dataset preparation, the molecules were read in Simplified Molecular Input Line Entry System
(SMILES) strings and transformed into molecular objects for graph representations in a 3D space for feature extraction and generating the adjacency matrix. 

First, the molecule was standardized, which included the following actions using RDKit: \texttt{RemoveHs()}, \texttt{SanitizeMol()}, \texttt{MetalDisconnector}, \texttt{Normalizer}, and \texttt{Reionizer}. Hydrogens were added to allow 3D optimization using the Merck Molecular Force Field (MMFF) \cite{halgren_merck_1996}. The original SMILES chirality was kept, if present. If the stereochemistry was not explicitly specified in the input SMILES string, it was inferred and assigned based on CIP (Cahn-Ingold-Prelog) rules using the 3D conformation of the molecule. Hydrogens were subsequently removed to prevent overexpression of specific properties and reduce data dilution. Finally, each node position is extracted from the 3D-generated molecule and used for feature extraction.

\subsection{Featuring Atoms, Bonds, and Molecules}
We selected chemical, physical, drug-like and spatial descriptors to enrich the topological information and aid in detecting patterns relevant to medicinal chemistry property predictions. The data values for each feature had mathematical operations tailored for each case, ensuring values ranging ~0-1, although batch normalization was later applied to ensure proper mathematical normalization per batch \cite{ioffe_batch_2015, lipinski_lead-_2004}. Table~\ref{tbl:feature_table} summarizes all selected features and operations made for each.

\begin{table}[h]
\centering
\small
\renewcommand{\arraystretch}{1.3}
\caption{Features for atoms, bonds, and global molecular properties}
\label{tbl:feature_table}
\begin{tabular}{ll}
    \hline
    \textbf{Atom Name} & \textbf{Atom Features} \\
    \hline
    Atomic Number & $(Z - 1) / 78$ \\
    Hybridization & sp, sp2 or sp3: [0, 0.5, 1] \\
    Electronegativity & $(\text{electronegativity} - 0.9) / 3.1$ \\
    Dipole Polarizability & $(\mathit{DP} - 4.5) / 31.5$ \\
    van der Waals Radius & $(\mathit{VdW} - 120) / 46$ \\
    Buried Volume & $\mathit{N_{\text{occupied}}}/{N_{\text{total}}}$ \\
    \hline
    \textbf{Bond Name} & \textbf{Bond Features} \\
    \hline
    Bond Length & Bond length $- 1$ \\
    Conjugated & [0, 1] \\
    Bond Type & Single, aromatic or double: $[1, 1.5, 2] / 2$ \\
    Ring Size & 0–8 $\rightarrow$ 0–1 \\
    \hline
    \textbf{Global Name} & \textbf{Global Features} \\
    \hline
    Chiral Centers & $\#\text{ chiral centers} / 6$ \\
    Hydrogen Balance & $\left(\mathit{HBD} / 5 - \mathit{HBA} / 10\right) / 10$ \\
    Rotatable Bonds & $\#\text{ Rotatable Bonds} / 10$ \\
    Solubility & $(\mathit{TPSA} + \mathit{LogP}) / 145$ \\
    SP3 Fraction & $\mathit{N_{\text{SP3}}} / \mathit{N_{\text{C}}}$ \\
    Radius of Gyration & Equation (19) \\
    \hline
\end{tabular}
\end{table}

The atomic numbers ($Z$) were min-max normalized in reference to hydrogen and platinum. The hybridization states were mapped to three values for sp, sp$_2$, and sp$_3$ with values 0, 0.5, and 1. The electronegativity values were min-max scaled using fluorine and francium as limits \cite{housecroft_inorganic_2018}. The dipole polarizability was scaled using hydrogen (4 Bohr$^3$) and iodine (35 Bohr$^3$) values. The van der Waals radius was normalized between hydrogen (1) and gold (79).

A 3D atomic descriptor, the buried volume, measures the occluded space within a spherical region ($R_i = 3.5$ \AA). This descriptor accounts for the steric effects that influence molecular interactions. The calculation involves generating a 3D grid with spacing $\lambda = 0.5$ \AA~and counting occupied nodes ($N_{\text{Occupied}}$) overlapping the van der Waals radius. The percentage of buried volume is computed using the equation in Table~\ref{tbl:feature_table}, generating a 3D spatial node feature.

The bond features include the length of the bond, obtained from optimized 3D structures using the Merck Forcefield \cite{halgren_merck_1996}, ensuring precision over empirical tables. The lengths were modified by subtracting 1 \AA. The conjugation status distinguishes double (1) and single bonds (0). Bond type values (single, aromatic, double as 1, 1.5, 2) were normalized by dividing over 2. The size of each ring (3-8 members) were mapped between values of 0 and 1.

Global molecular features include chiral centers, scaled by simply dividing over 6, based on MoleculeNet\cite{wu_moleculenet_2018} distributions where more than six chiral centers were rare (Supplemental Information, Appendix A, A)). Hydrogen bond donors (HBD) and acceptors (HBA) form the "Hydrogen Balance" function explicit in Table~\ref{tbl:feature_table}, yielding values of 0.2-1 for drug-like molecules and $<$0.2 for Lipinski violations. A safeguard replaces zero denominators with $10^{-10}$. 

The number of rotatable bonds were divided over 10, based on the Veber’s rule (max = 10) \cite{veber_molecular_2002}. The solubility feature combines logP and the topological polar surface area (TPSA), preserving their complementary roles in predicting permeability and bioavailability. Given TPSA $\leq 140$ \AA$^2$ and logP $<$ 5 for oral drugs, their sum (145) was used as normalization factor \cite{mobitz_design_2024}. The fraction of sp$^3$ carbons quantifies the fraction of carbons in sp3 hybridization per molecule.

Finally, we chose to work with the molecular radius of gyration ($r_{RG}$) because of its capability to describe the spatial distribution of each molecule, It calculation corresponds to the mass-weighted root-mean-square (RMS) deviation from the center of mass, given by ($r_{CM}$):

\begin{equation}
    r_{RG} = \sqrt{\frac{\sum_{i=1}^{N} m_i (r_i - r_{CM})^2}{\sum_{i=1}^{N} m_i}}, \quad r_{CM} = \frac{\sum_{i=1}^{N} m_i r_i}{\sum_{i=1}^{N} m_i}\tag{20}
\end{equation}

\subsection{Dataset Preparation}

The architecture models were written in pytorch.geometric \cite{fey_fast_2019}. For compatibility with pytorch, each molecule was transformed into a tensor-like object containing atom-level ($x$), bond-level ($\text{edge\_index}$, $\text{edge\_attr}$), and global ($u$) features using the \textit{DataLoader} function. Labels ($y$) were optional for prediction tasks but required for supervised learning. 

The dataset class also retains metadata such as molecule names and SMILES strings for traceability and debugging. During debugging, we identified the terms ``.[H+].[Cl-]'' in instances of the BACE dataset that were eliminated to bypass errors during data processing.

The class imbalance (0 to 1 ratio in binary classes) for the BACE dataset was 1.2:1 and  0.31:1 for the BBBP. For the TRPA1, we set a threshold of 100 nM to distinguish actives from inactives, this limit yielded an imbalance of 0.63:1. In the case of the Lipophilicity dataset the majority of the logD values were in the range of 2.7-3.3 (total range is -1.5 to 4.5) We noticed that several repeated SMILES were found in the TRPA1 dataset that corresponded to same molecules with different reads of IC50 values, we kept them all for reproducibility purposes.
To handle class imbalance, we generated alternative SMILES to compensate for the less represented class in the BACE and BBBP datasets. Additionally, we used the \textit{weighted random sampling} function to assign each label a probability for compounds to be selected during training. Weights were calculated using the reciprocal of their corresponding class count \cite{efraimidis_weighted_2008}.

\subsection{GNN Trainer Functionalization}
The training method executes multiple forward and backward passes over the dataset for 50 epochs, producing global binary predictions or scalar values depending on the task specified: Classification or Regression. Optionally, node-level outputs can be visualized as colormaps projected onto molecular images.

The training process employs the Adam optimizer \cite{kingma_adam_2017} to update model parameters. A learning rate scheduler, ReduceLROnPlateau, dynamically adjusts the learning rate. The loss is computed using (\textit{BCEWithLogitsLoss}) in the case of classification and  (\textit{MSELoss}) for regression. Backpropagation optimizes embeddings, incorporating gradient clipping ($\text{max norm} = 1$) via $\text{torch.nn.utils.clip\_grad\_norm}$.

Each batch is processed independently using $\text{optimizer.zero\_grad}$ to clear previous gradients. The loss is tracked and plotted per batch, weighted by the number of graphs, and accumulated into the total loss. The method ultimately returns the average loss per graph for the entire dataset.

Performance evaluation for classification includes the F1 score, accuracy, and area under the receiver operating characteristic curve (ROC-AUC) derived from a True-Positive-Rate (TPR) vs. False-Positive-Rate (FPR) plot:

\begin{equation}
\mathrm{AUC} = \int_{0}^{1} \mathrm{TPR}(\mathrm{FPR}) \, d(\mathrm{FPR})\tag{21}
\end{equation}

The F1 score is computed using the True Predictions (TP) and the False Negatives (FN):
\begin{equation}
    F_1 = \frac{TP}{TP + \frac{1}{2}(FP + FN)}\tag{22}
\end{equation}
Accuracy is calculated using the same inputs as F1 and adding True Negatives (TN) and False Positives (FP):

\begin{equation}
    \text{Accuracy} = \frac{TP + TN}{TP + TN + FP + FN}\tag{23}
\end{equation}

For assessing the error of prediction in the regression task, we use the Root-Mean-Square Error to calculate the error between predictions $y_{\text{pred},i}$, and actual values $y_{\text{true},i}$: 

\begin{equation}
\mathrm{RMSE} = \sqrt{\frac{1}{N} \sum_{i=1}^{N} \left( y_{\text{true},i} - y_{\text{pred},i} \right)^2}\tag{24}
\end{equation}

These metrics estimate predictive performance per epoch during training and evaluation.

\subsection{Feature Selection and Hyperparameter Optimization}

Feature selection employed the BMP architecture with predefined hyperparameters: hidden channels = 250, learning rate = 0.003, batch size = 32, epochs = 50, dropout rate = 0.25. 

To identify the most relevant features based on validation F1 score contributions, we employed a hybrid backward/forward elimination which combines all 16 initial features with the resulting F1 score, serving as cut-off for subsequent rounds of elimination. Features whose removal improves the initial cut-off are eliminated. A mini-forward selection adjustment was applied to confirm optimal F1 score improvement. From this, features were re-added from the elimination list to assess their impact. Features that did not enhance the model scoring were removed. This process was repeated until no further improvement was observed.

After feature selection, hyperparameter tuning was performed separately for each of the five models on a per-dataset basis, resulting in a total of 15 optimization procedures. The hyperparameter protocol was carried out using the \textit{Optuna} package \cite{akiba_optuna_2019}, the optimized function runs a cross-validation protocol and works in a dual direction, including the minimization of the loss difference between validation and training to fight overfitting and the maximization of the validation F1-score. The exploration of the hyperparameter space utilized the Tree-Structured Parzen Estimator (TPESampler) \cite{bergstra_algorithms_2011}. TPE is a Bayesian method that efficiently navigates the hyperparameter space by adapting modeling the likelihood of achieving a high-performing configuration based on previous trials. To accelerate the process, we introduced a pruning protocol; after 30 epochs, if the validation F1 score is below 0.65 the trial is terminated. 

The optimized parameters were the number of hidden channels, batch size, and the dropout rate. 

\subsection{Depiction of Atom Relevance in Predictions}

To analyze relevance across atoms during the node-block phase, the rdkit.MolDraw2DCairo package was used for projecting predictions onto molecular structure images. The generation of the node-importance scores involved a linear transformation that mapped the outcoming channels from previous processing out to a single channel, followed by a sigmoid activation function to produce scalar values as scores. These scores quantify the contribution of each node to the global prediction task. Since the ongoing embeddings are extracted before the global pooling operation, they capture a snapshot of the message passed onto each atom, providing insights into the passed message.

The generation of the node-relevance images starts with a SMILES to 2D transformation adhering to how the molecules were transformed into graphs for training, this ensures obtaining the same stereochemistry and number of atoms as the node embeddings resulted for each graph. Since the output from the model is in the form of an array per batch processed, a tracking of the number of atoms per graph helps to slice the embedding per molecule. Finally, atom importance values are normalized on a molecule basis using min-max scaling. 

\section{Results and Discussion}

This study evaluates the binary prediction performance of five distinct message-passing-based models on three datasets characterized by message decomposition, integration of attention and convolution factors and approaches to achieve undirected graphs. To ensure a fair comparison, we optimized the selection of features for each dataset and tuned hyperparameters for each model.

The dataset preparation code efficiently processes molecules and excludes failed SMILES entries, while also logging where each failure occurred to support debugging.

Computation time was a key consideration during code development, as most drug discovery campaigns require predictions for thousands to millions of compounds. To offer a sense of computational efficiency, we performed one round of training on approximately 3,000 molecules for 150 epochs, followed by prediction on ~1,500 compounds. The entire process completed in 14 minutes on a single NVIDIA GeForce RTX 4070 GPU.

Below, we present and discuss the results of the optimization process, including both feature selection and hyperparameter tuning. We then compare the performance of all tested models internally, evaluating both global and node-level predictions. Finally, we benchmark our best-performing models against widely used machine learning methods in MoleculeNet datasets.

\subsection{Feature selection and the effect of 3D conformations in predictive performance}

The feature selection process was the first step towards adapting our models for each dataset. We chose the least count of features to reduce noise or redundant information \cite{cai_feature_2018} implementing an iterative hybrid feature selection approach that combined backward elimination with a forward selection process for refinement.

In each iteration, we temporarily removed one feature at a time and ranked all features based on their validation F1 scores: the feature whose removal caused the largest drop in F1 score received the highest number of points, based on rank positions, while those whose removal improved performance were ranked lower. From this ranking, we identified a candidate subset of features whose removal improved the F1 score compared to the baseline (using all features in the round considered). 

\begin{figure}[h]
\centering
  \includegraphics[height=8.8cm]{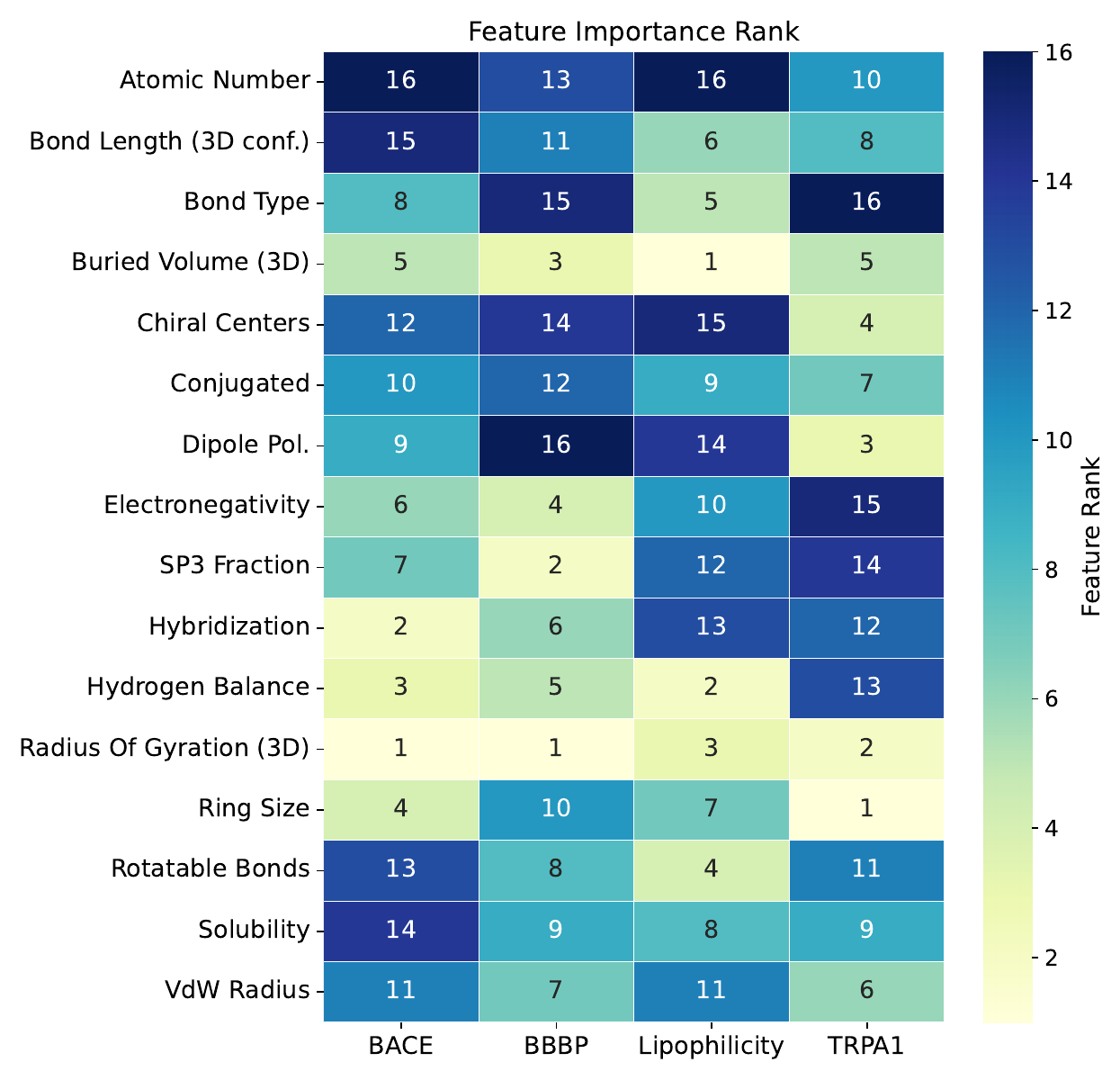}
  \caption{ Ranked features based on the cumulative sum of their positions across successive rounds of backward elimination. In each round, features were ordered from lower to higher F1 score upon removal, and points were assigned accordingly. Final ranks were determined by the total accumulated points, with higher scores indicating greater importance. The heatmap visually highlights the consistency of feature rankings across datasets.
}
  \label{fgr:ranks}
\end{figure}

We then applied a forward-like refinement process, adding features from the elimination list one by one. At the end of the selection process, the feature with the largest cumulative of points in the rounds was ranked at the top. In this manner, we are able to rank the features that consistently benefited learning for each dataset across all rounds, the results ranking tables are summarized in Fig.~\ref{fgr:ranks}. 

The feature importance rank indicates that the Buried Volume and Radius of Gyration, a node and global 3D features, consistently achieved 1-5 places across all datasets. Ring size, a 2D edge feature, also plays an important role in the prediction of binding activity. In contrast, the atomic number, chiral centers, conjugated, solubility and VdW radius had the lowest importance ranks across the datasets.

Moreover, the list of eliminated features and the final metrics are summarized in Table~\ref{table:feature_selection}. The starting reference F1 values for the 16 features were 0.82, 0.92 and 0.80 for the BACE, BBBP and TRPA1 datasets, respectively. While the validation RMSE value for the 16 features included in the case of the Lipophilicity was 0.866. We identified that some of the features of the elimination list are correlated with each other. For example, in the case of BBBP, solubility was correlated with H balance (0.83), atomic number with electronegativity (0.98) and conjugation feature with bond type (0.75) and bond length (0.77). Either one of these was eliminated in the BACE and BBBP and all of these in the TRPA1. This redundancy explains why these features were discarded during feature selection.
\begin{table}[h]
\centering
\small
\captionsetup{skip=8pt} 
\caption{Summary of eliminated features per dataset and obtained validation F1 score during cross-validation.}
\label{table:feature_selection}
\renewcommand{\arraystretch}{1.2}
\begin{tabular}{l p{0.4\textwidth} c c}
    \hline
    \textbf{Dataset} & \textbf{Eliminated Features} & \textbf{Val.} & \textbf{No 3D} \\
    \hline
    BACE  & Rot. Bonds, Solub., Atomic N., Bond L., Conjug. & F1: 0.83 & F1: 0.77 \\
    BBBP  & H-Bal., Electro.                               & F1: 0.93 & F1: 0.78 \\
    TRPA1 & Frac. SP3, Electro., Conjug., Rot. Bonds, Atomic N., H-Bal., Hybrid., Solub. 
          & F1: 0.84 & F1: 0.78 \\
    Lipo. & Atomic N.                                      & RMSE: 0.65 & RMSE: 1.14 \\
    \hline
\end{tabular}
\end{table}

\begin{figure}[h]
\centering
  \includegraphics[height=13cm]{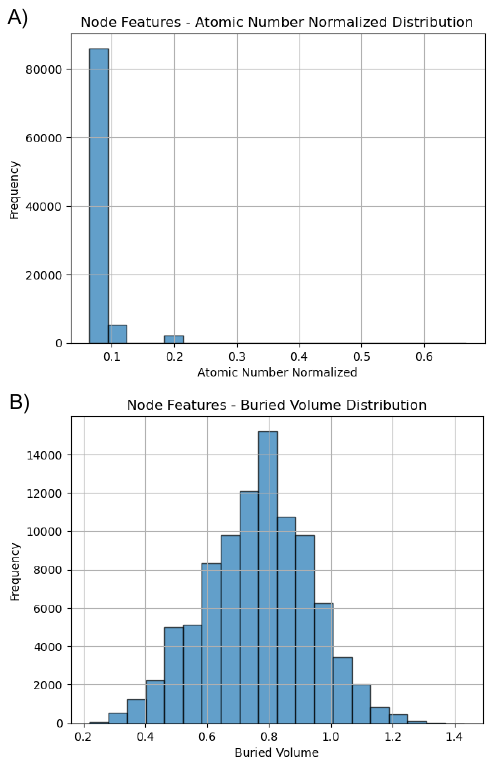}
  \caption{ Frequency histograms comparing single-valued features per element with 3D features that are 3D-enviornment aware working with the TRPA1 set. In A) the distribution for the normalized atomic number, the highest frequency count corresponds to the carbon element. Hydrogens were eliminated during data processing.B) The standardized buried volume node feature distribution shows a gaussian-like distribution.
}
  \label{fgr:3}
\end{figure}

Another possible explanation is an inherent problem when working with element-like features, such as the atomic number or the VdW radius, to represent organic compounds: they contain more carbons than any other type of element. Hence, using features that are invariant to the molecular context suffers from a skewed distribution. To address this data limitation, we propose incorporating environment-dependent 3D features, such as the buried volume and radius of gyration, which were top-ranked across all datasets. These features exhibit Gaussian-like distributions (Fig. \ref{fgr:3}), which is desirable during training because it helps prevent model bias and mitigates oversmoothing. In addition, Table \ref{table:feature_selection} shows that eliminating these 3D features along with the bond length reduces the F1 score by 6-15\%, which is a significant drop in the context of drug discovery applications. For example, a screening pipeline that evaluates 1 million drug candidates could result in approximately 60-150,000 compounds being misclassified. These findings highlight the critical role of spatial features in capturing meaningful structural variation, which ultimately increases the predictive performance of MPNN models for large-scale drug discovery efforts.

\begin{figure}[H]
\centering
  \includegraphics[height=7.5cm]{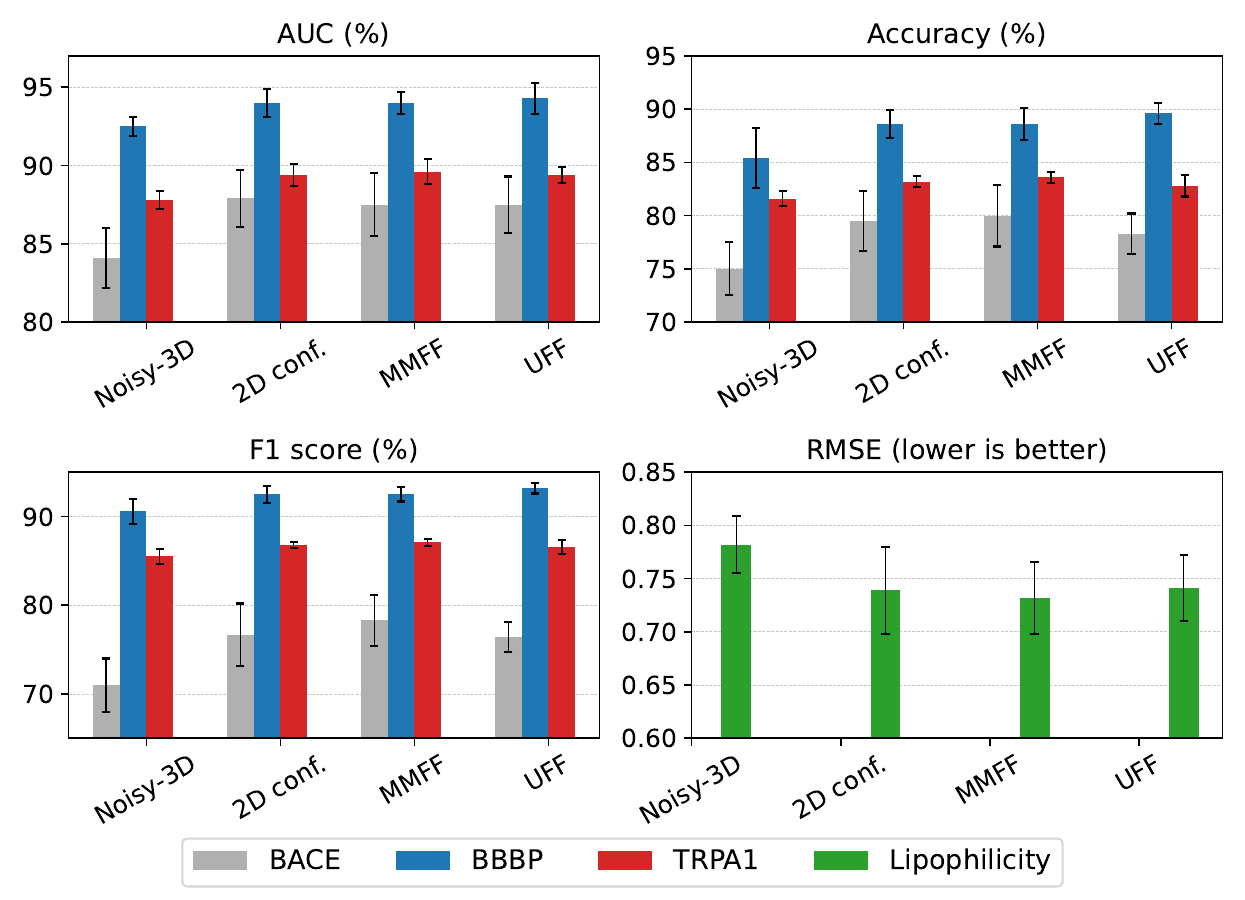}
  \caption{ Ablation study to address the influence of the spatial arrangement of the molecules across the tested datasets (BACE, BBBP, TRPA1 and Lipophilicity). We included 3D conformations with added gaussian noise (0.5 Å std. dev.) termed Noisy-3D, 2D conformations only and spatial optimizations using the Merck Molecular Force-Field (MMFF) and the Universal Force-Field (UFF). The plots are separated by classification metrics (AUC, Accuracy, F1) and regression (RMSE). Error bars represent margins of errors (95\% confidence) over multiple runs.
}
  \label{fgr:5}
\end{figure}

We observed that incorporating 3D-derived features improves predictive performance; however, generating accurate 3D molecular conformations is computationally expensive and time consuming. To better understand the necessity of 3D spatial information for predictive modeling, we conducted an ablation study using the four datasets employed in this work Fig.~\ref{fgr:5}.

In this study, we compared several scenarios: (1) models using only 2D molecular representations with derived 3D features, (2) models using perturbed 3D conformations created by adding Gaussian noise (0.5 Å std. dev.) to each atomic (X, Y, Z) coordinate, and (3) models using optimized 3D conformations generated with two force fields, the Merck Molecular Force Field, designed and applied for a wide range of organic molecules (MMFF)~\cite{halgren_merck_1996}, and the Universal Force Field(UFF), a full periodic table covered force field ~\cite{rappe_uff_1992}. 

This approach allowed us to assess the importance of accurate 3D conformations in predictive performance. The results for all datasets demonstrate that Noisy-3D structures, significantly degraded performance, indicating that inaccurate 3D geometries can be detrimental. 

Interestingly, the predictive metrics were comparable between models using only 2D features and those using optimized 3D conformations (MMFF or UFF), suggesting that incorporating 3D-derived features from 2D representations may be sufficient for robust prediction. Furthermore, there was not meaningful difference between the MMFF and the UFF force fields for optimizing 3D configurations. Under this scenario the dataset preparation for 2D molecules is 2.3 times faster compared to including the MMFF based spatial optimization, underscoring that 2D configurations not only keep predictive accuracy intact but also significantly speed up computation, making them a more practical and scalable option for high-throughput screening.

Another aspect of the feature selection results is that the number of total features was significantly lower for the TRPA1 dataset (8), followed by BACE (11), BBBP (14) and Lipophilicity (15). Such behavior led us to hypothesize that underlying variety in the chemical space might be influencing the dependency on input features. In this sense, datasets with different chemotypes could require a wider feature set to capture the range of variations.

Diversity is defined as to certain properties of a system that contains items that are classified into types. Specifically, these properties include the number of types, the way items are assigned to those types, and the different types from each other. In molecular analysis, this results in the classification of chemical compounds (items) into chemotypes (types) based on their structural features~\cite{Morales_diversity}. The diversity of a chemical library can thus be analyzed with structural cluster analysis to obtain how many chemotypes are present, via the number of clusters, and how compounds are distributed among them through Shannon entropy~\cite{shannon1948mathematical}. 

We performed the structural clustering using the BitBIRCH algorithm with a similarity threshold of 70\% (Diameter with pruning mode, tolerance = 0.05)\cite{perez_bitbirch_2025} working with RDKit fingerprints (\textit{nBits} = 2048)~\cite{landrum_rdkitrdkit_2025}. 

Based on the results (Table ~\ref{table:dataset_statistics}), where a higher number of clusters and Shannon entropy means higher diversity, there appears to be a correlation between diversity and the number of features required to represent the data, confirming our hypothesis: greater chemical diversity, exemplified with the Lipophilicity case containing 52\% clusters that were singletons, would require more features, 15 out of 16 in this case, to help the model learn the wide range of chemotypes within classes. 
\begin{table}[h]
\centering
\small
\caption{Summary of dataset statistics and selected features after feature elimination.}
\label{table:dataset_statistics}
\renewcommand{\arraystretch}{1.2}
\begin{tabular}{lrrrr}
    \hline
    \textbf{Dataset} & \textbf{Size} & \textbf{Clusters} & \textbf{Shannon E.} & \textbf{\# Features} \\
    \hline
    Lipo. & 4200 & 2642 & 10.74 & 15 \\
    BBBP  & 2038 & 1143 &  9.46 & 14 \\
    BACE  & 1513 &  186 &  6.14 & 11 \\
    TRPA1 & 3014 &  305 &  6.03 &  8 \\
    \hline
\end{tabular}
\end{table}

\subsection{Hyperparameter Optimization}

After selecting the most relevant features, we proceeded with a dual-goal process of hyperparameter tuning to refine model performance utilizing 5-fold cross-validation. The optimization strategy had a dual direction in the case of the classification task: the first, minimizing the difference between validation and training loss to address overfitting, a pervasive issue found in this study; second, we sought to maximize the F1 score, which was selected due to its emphasis on positive label identification.

For optimizing hyperparameters in the case of regression we minimized both the difference of the validation and the training loss and the validation RMSE loss.

\begin{figure*}[h]
 \centering
  \includegraphics[width=0.98\textwidth]{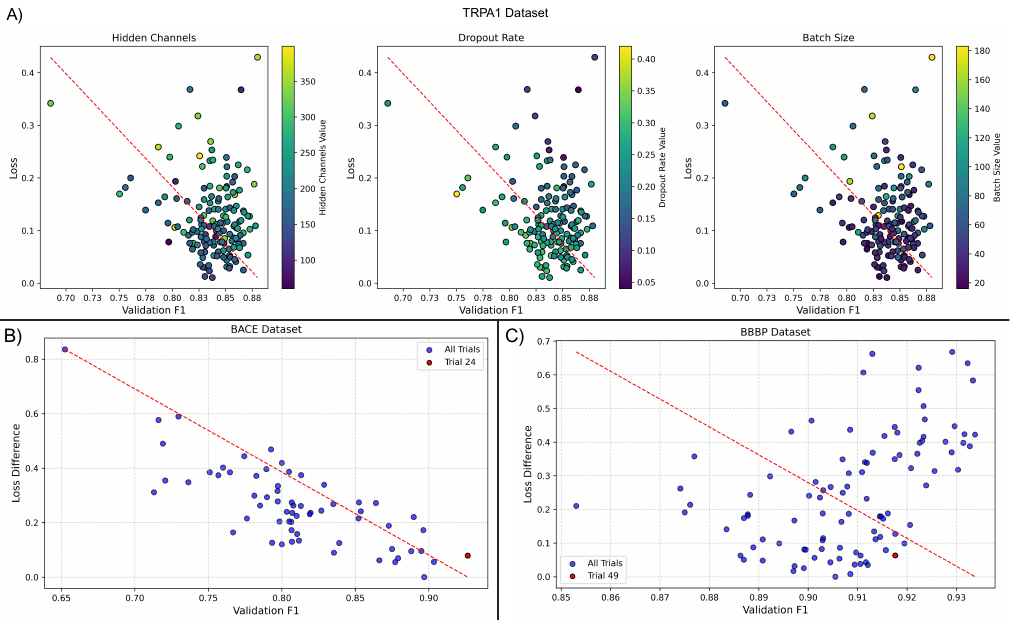}
  \caption{Pareto plots for the BMP model showing the convergence of the dual-directed hyperparameter optimization protocol using TPE sampler and run using the Optuna package.
  \textbf{A)} TRPA1 dataset panel visualizing the relationships between the four features optimized, hidden channel number (50-400), dropout date (0.05-0.5), batch size (20-180) with indicated ranges of optimization in the colormap bars, the closer to yellow colors the higher the feature value.
  \textbf{B)} Pareto plot for the BACE dataset showing two colors, the blue dots indicates tested trials, while the the red one corresponds to the trial we selected for our final models. 
  \textbf{C)} Pareto plot for the BBBP dataset, the optimization did not converged into the dual-direction minimization, rather a trade-off between the two directions is observed.}
  \label{fgr:6}
\end{figure*}

The optimization protocol effectively identified the ideal parameter configurations for most models, demonstrating the utility of the protocol (Supplementary Information, Appendix C). Convergence was achieved for the four parameters of the TRPA1 after 200 trials and only a 100 for the BACE case (Fig.~\ref{fgr:4}, section A and B). The dual-optimization protocol had difficulty converging to values that balanced both objectives for the BBBP dataset, as seen in a linear tendency to trade F1 for an increased loss difference (Fig.~\ref{fgr:6}, section C). In the lipophilicity trials, the model consistently achieved low loss values with acceptable RMSE values with relative ease. However, an issue arose wherein negative loss difference values were observed. To mitigate this, we implemented a pruning criterion whereby, after 20 epochs, any trial exhibiting an absolute loss difference greater than 0.15 was discarded.

The non-convergence observed for the BBBP dataset may be due to its poor diversity. In this scenario, there is a higher risk for overfitting. Because of this reason, we gave preference for a lower loss difference rather than the original dual direction for optimization. 

\subsection{Testing directionality, self-node, convolution and attention Mechanism in the message-passing architecture}

We evaluated five MPNN variants with modifications in the node block. Each variation incrementally increased complexity relative to the baseline model (BMP) but remained less complex than the reference model, UMP. Model complexity, quantified by parameter counts, is shown in Table~\ref{table:model_summary}. Starting from BMP, which lacks node self-perception, we tested  incorporating self-nodes, attention, and convolution. The UMP model, used as a reference and containing the higher count of parameters, operates on undirected graphs achieved by duplicating node connections in the adjacency matrix and includes self-node processing \cite{battaglia_relational_2018}.

\begin{table}[h]
\centering
\small
\caption{Parameter counts per node and total (which includes 68,000 weights for the message block and 128,521 for the global block). Training time for the model architectures on the BACE dataset is also reported.}
\label{table:model_summary}
\renewcommand{\arraystretch}{1.2} 
\begin{tabular}{lccc}
    \hline
    \textbf{Model} & \textbf{Node} & \textbf{Total} & \textbf{Time (s)} \\
    \hline
    BMP       & 189,251 & 385,502 & 8.39 \\
    CBMP      & 189,251 & 385,502 & 8.97 \\
    BMP+SN    & 190,751 & 387,002 & 8.52 \\
    ABMP      & 193,501 & 389,752 & 9.99 \\
    ABMP+SN   & 195,001 & 391,252 & 9.57 \\
    UMP       & 256,251 & 452,502 & 9.85 \\
    \hline
\end{tabular}
\end{table}

\begin{figure}[H]
\centering
  \includegraphics[height=15cm]{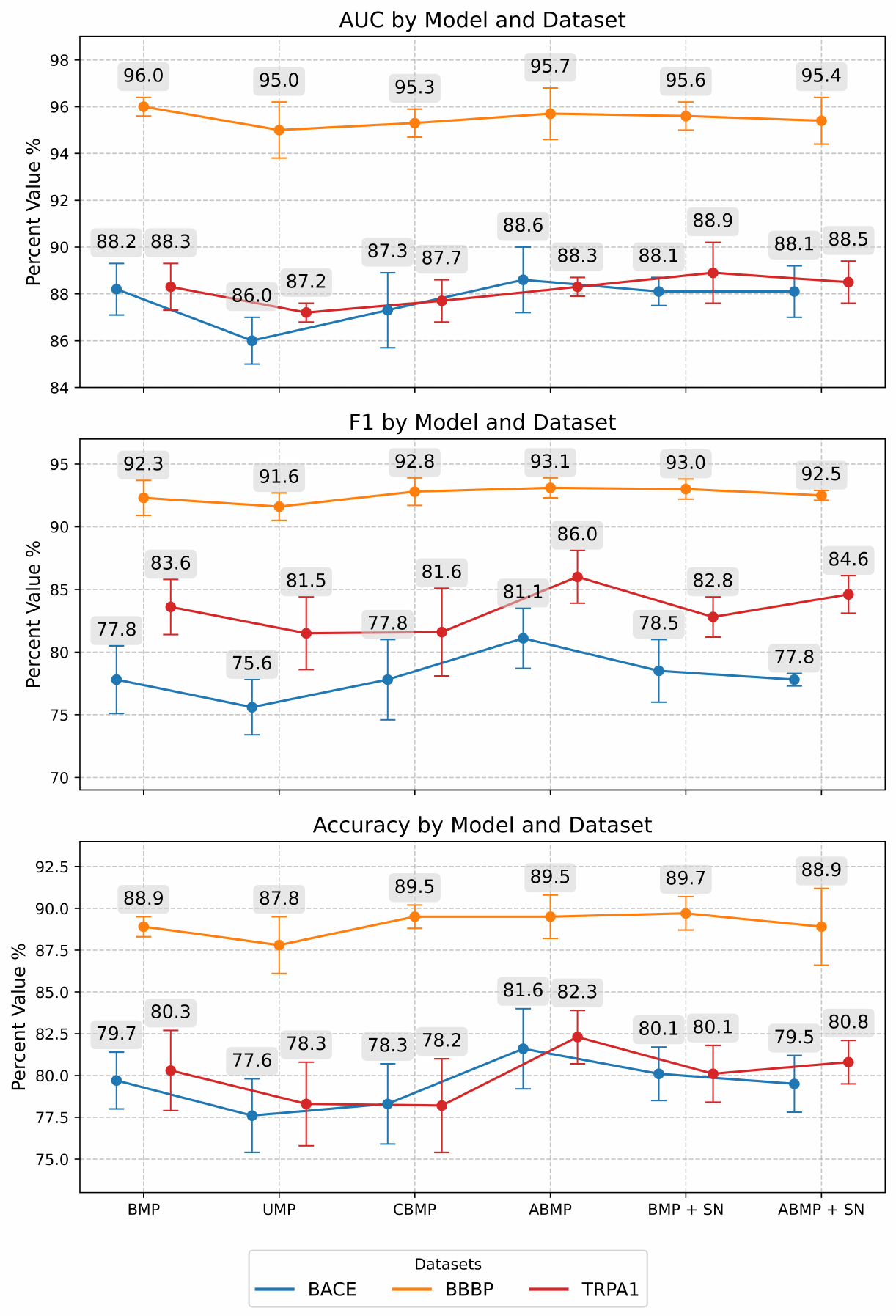}
  \caption{ Performance comparison of the five models across three datasets (BACE, BBBP, TRPA1). These values are averaged using five random seeds on blind test sets (20\% of the total dataset). Metrics shown include AUC (top), F1 score (middle), and accuracy (bottom) with error bars indicating the margin of error with a confidence of 95\%. The datasets are color-coded for clarity, and the x-axis represents the model indices. 
}
  \label{fgr:7}
\end{figure}

The ABMP + SN architecture contains attention and self-perception components added to the BMP. The overall results for classification tasks are displayed in Fig.~\ref{fgr:7} and include the average AUC, F1 and accuracy metrics values on blind tests (20\% of the total dataset) averaged across five different random seeds using the optimized hyperparameters for each model and dataset. 

In general, the metrics have subtle differences within each dataset for the classification tasks. In the case of the BBBP dataset, either method achieved high performance in F1 and AUC scores specifically across all models. The BBBP data set exhibits a strong class imbalance, with the positive label being overrepresented by 82\% according to the average class proportion between clusters (Figure 2. \textbf{B)} in \textit{Supplemental Information}) which explains why the accuracy values were not as high as the F1 scores. In adittion, the TRPA1 dataset has higher F1-scores compared to BACE dataset, this is because the class imbalance favors the positive class for the TRPA1 case, making it easier for any model to learn on this class. Since the F1 score focuses on measuring how well the model is doing at identifying actives, it was expected to obtain higher F1 scores than BACE dataset.

In an attempt to balance classes, we tested whether duplicating the less represented class would have a positive impact on performance. To do this, we exploited the fact that SMILES strings can have multiple valid representations along with the weighted random sampling function, full protocol detail can be found in \textit{Supplemental Information Appendix B}. Based on the results of the confusion matrix, we found that this duplication strategy is effective only when the primary objective is to improve the prediction of the majority class, producing higher F1 scores, but this comes at the expense of reduced accuracy for misrepresented labels.

We previously assessed intra-dataset diversity, identifying BBBP as more diverse than TRPA1 and BACE. However, the distinctiveness of chemotypes within each dataset likely also influences the ability of the model to distinguish between classes. The higher performance observed on BBBP may be attributed to greater structural dissimilarity among its chemotypes, which facilitates learning discriminative features.

\begin{figure}[h]
\centering
  \includegraphics[height=8.5cm]{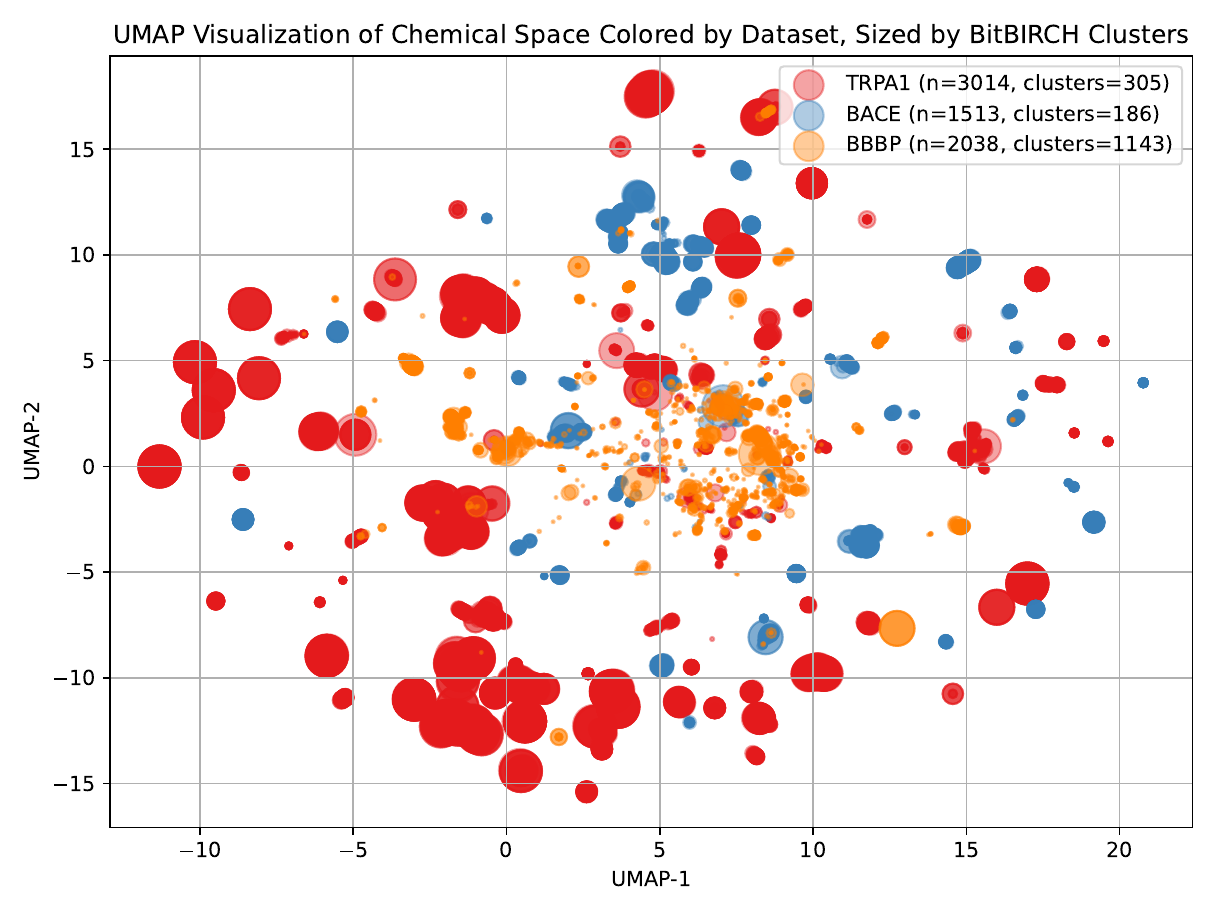}
  \caption{UMAP projection of RDKit fingerprints (2048 bits) for three datasets (TRPA1, BACE, and BBBP), illustrating their respective chemical space distributions. Each point represents a compound, colored by dataset, where blue, orange and green correspond to TRPA1, BACE and BBBP datasets respectively. Each points was re-dimensioned to represent their respective BitBIRCH cluster sizes. The visualization highlights the poor structural diversity within the BBBP case, while TRPA1 shows more spread and well-formed clusters. 
}
  \label{fgr:4}
\end{figure}

To evaluate this, we applied Uniform Manifold Approximation and Projection (UMAP) to reduce the high-dimensional chemical space into a two-dimensional representation, enabling visual inspection of structural relationships between chemotypes (Figure~\ref{fgr:4}). The projection was generated using a globally focused configuration with parameters $\textit{n\_neighbors} = 50$ and $\textit{min\_dist} = 0.1$, employing the Jaccard distance metric to capture the dissimilarities of molecular fingerprints.

Among the classification datasets, the BBBP dataset represents an extreme case, exhibiting the lowest structural dissimilarity despite its relatively high number of clusters and singletons. This is evident in the UMAP projection, which shows a more compact distribution across both dimensions compared to the other datasets.

Interestingly, the simplest architecture tested, the BMP, with the lowest parameter count, achieves the highest AUC scores with low variance on this dataset. This suggests a relationship between model complexity and dataset structural dissimilarity: increasing architectural complexity does not necessarily improve learning performance. Rather, when structural dissimilarity is low, limiting the amount of information introduced during message passing appears to promote both model stability and sufficient representational power.

The model complexity analysis for small graphs, such as molecules, provides insights into just how simple a model can be while still performing well. Our findings support the hypothesis that redundant feature amplification may be unnecessary for small graphs, like molecules, which typically contain only 20–70 heavy atoms \cite{isert_qmugs_2022}. Moreover, the upper bound of effective complexity seems to be illustrated by the fact that combining two top performing models (ABMP + SN) did not yield significant improvement over their individual components (ABMP and BMP+SN), further suggesting that added architectural complexity may not be beneficial for small, structurally simple graphs.

\textbf{In the CBMP,} the use of convolution does not improve performance on any of the models, discouraging the use of this strategy in bidirectional MPNNs. Convolutional normalization operates by penalizing nodes with higher degrees of connectivity. However, this step appears unnecessary for applications involving organic molecules, where the connectivity degree typically ranges from 1 to 4. Moreover, degree imbalance has already been addressed by removing hydrogen atoms during graph pre-processing. Removing hydrogen atoms from molecular graphs yields a degree distribution that is notably closer to a Gaussian shape, as evidenced by a substantial reduction in skewness, from 0.321 to -0.016, and an increase in kurtosis, from 1.493 to 2.494, working with the Lipophilicity dataset as example (see atom-degree distribution in Supplementary Information, Appendix A, Figure 2).

\textbf{Comparing BMPs to UMP, } the BMPs scores were consistently higher in all metrics compared to their parent seed, the UMP. We investigated if the pooling type could play a factor. We replaced the mean-pooling operation with max-pooling and evaluated the model on the BACE dataset. The results were comparable to those of the BMP models (UMP with max-pooling using BACE dataset: AUC = $88.0~\pm~0.8$, F1 = $79.4~\pm~1.4$, Accuracy = $80.7~\pm~1.4$ ). One possible explanation is that max-pooling enhances class separability by emphasizing dominant node or edge embeddings. In this case, combining a synthetically enlarged adjacency matrix with a mean pooling operation would dilute the data since we are averaging over a larger tensor. This offers an explanation to the lower performance of the UMP, where irrelevant or noisy signals are retained rather than filtered out when using max-pooling. 
Noteworthy, the UMP architecture uses two MLP transformations in the node block, while the BMPs use only one, yet the latter achieves same scoring metrics highlighting that reduced data processing could yield similar results. Overall, our results suggest that bidirectional message-passing combined with max-pooling yields stronger representations and that reduced data processing achieves similar results to the UMP.

\textbf{The ABMP model architecture, } obtained the highest scores in all metrics for TRPA1 and BACE indicating that our attention mechanism benefits the bidirectional message-passing architecture.

\begin{figure}[h]
\centering
  \includegraphics[height=14cm]{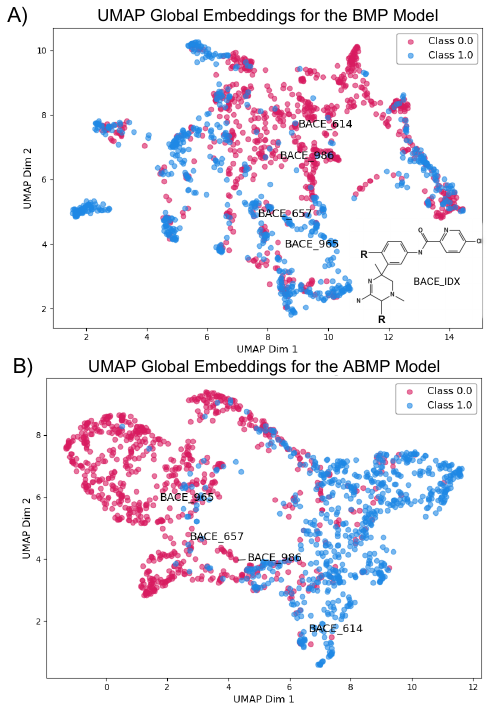}
  \caption{ Non-linear dimensionality reduction analysis (UMAP) for the BMP and ABMP Global Embeddings of the BACE dataset. Red dots belong to the class 0, representing the embeddings of BACE inhibitors, while in blue are the non-inhibitors embeddings. BACE\_614 and BACE\_657 are both active, while BACE\_657 and BACE\_965 and BACE\_986 are inactive. \textbf{A)} UMAP Global embeddings for BMP. \textbf{B)} UMAP Global embeddings for ABMP. 
}
  \label{fgr:8}
\end{figure}

To analyze the ABMP classification capacity, we generated a non-linear dimensionality reduction analysis using UMAP to visualize the ability to discern between classes of these two specific models. We projected the global-output embeddings with highlighted classes in blue (1) and red (0) and included the identifiers of four molecules with a common scaffold to analyze if these models are capable of distinguishing them by classes (Fig.~\ref{fgr:8}). 

There is more class separation in the ABMP model compared to the BMP, as observed in the UMAPs, suggesting that the attention mechanism enriches the representations with information useful for label separability. To ensure this improvement was attributable to the attention mechanism itself rather than any unrelated mathematical artifact, we applied the same attention layer to three different base models: a single forward-directed message-passing (MP), the UMP, and the BMP (Table ~\ref{table:attention_direction}).

\begin{table}[h]
\centering
\small
\caption{Ablation study analyzing the influence of the attention mechanism in message passing with single, bidirected, and undirected (duplicated adjacency matrix) graphs. The dataset used was BACE.}
\label{table:attention_direction}
\renewcommand{\arraystretch}{1.2}
\begin{tabular}{lllccc}
    \hline
    \textbf{Attn.} & \textbf{Dir.} & \textbf{Model} & \textbf{F1 (\%)} & \textbf{Acc. (\%)} & \textbf{AUC (\%)} \\
    \hline
    (+) & Bi-dir. & ABMP  & 81.1 $\pm$ 2.4 & 81.6 $\pm$ 2.4 & 88.6 $\pm$ 1.4 \\
    (-) & Bi-dir. & BMP   & 77.8 $\pm$ 2.7 & 79.7 $\pm$ 1.7 & 88.2 $\pm$ 1.1 \\
    (+) & Single  & AMP   & 71.0 $\pm$ 3.1 & 76.3 $\pm$ 1.9 & 85.6 $\pm$ 1.2 \\
    (-) & Single  & MP    & 61.9 $\pm$ 2.5 & 72.2 $\pm$ 0.8 & 83.5 $\pm$ 2.3 \\
    (+) & Undir.  & AUMP  & 77.5 $\pm$ 2.2 & 79.7 $\pm$ 2.6 & 87.3 $\pm$ 1.9 \\
    (-) & Undir.  & UMP   & 75.6 $\pm$ 2.2 & 77.6 $\pm$ 2.2 & 86.0 $\pm$ 1.0 \\
    \hline
\end{tabular}
\end{table}

In the three cases tested the attention mechanism increase predictive scores compared to their parent references, underscoring that it is indeed the attention mechanism which enhances performance on the message-passing frameworks studied. This naturally leads to the question: \textbf{how does attention reshape message-passing to better capture relevant chemical regions for the property prediction?} We identify three main distinctions from classical GAT that explain this behavior:

First, we utilize edge-aware attention, which is wise since bonds (edges) act as natural importance flags, by explicitly encoding the relationship between a pair of atoms. Perhaps a very intuitive example of this is the conjugation state, which acts as an electronic gate that dictates $\pi$-electrons delocalization. Moreover, bond features are interatomic descriptors that allow us to describe many different aspects in one single value that can be relevant for the property prediction task. For example, the bond length is correlated with torsional angles and resonance. By explicitly integrating these edge-derived signals, the attention mechanism is better equipped to capture how such subtle structural-electronic features between atoms govern macroscopic molecular properties.

Second, our ABMP model applies independent linear transformations to the edge, source, and target features before combining them into an attention coefficient that acts as an importance factor directly onto the message. By first processing molecular components (bonds concatenated to source and target nodes) independently within the attention mechanism, we exploit the information available in each component prior to integration. This design is particularly suitable for message-level attention, as all components that constitute the message are explicitly considered in the attention computation, ensuring that the coefficient reflects the full context of the message it is acting on.

Third, instead of summing attention-weighted messages as in classical GAT, we use scatter-max pooling, which mitigates the cardinality problem observed in GATs, where high-degree nodes are overrepresented when neighbor embeddings are summed~\cite{zhang_improving_2020}. While this issue is less pronounced when working with organic molecules, it can still bias attention in standard GAT formulations. Scatter-max instead emphasizes the strongest, most predictive neighbor messages. 

\begin{figure}[h]
\centering
  \includegraphics[height=8cm]{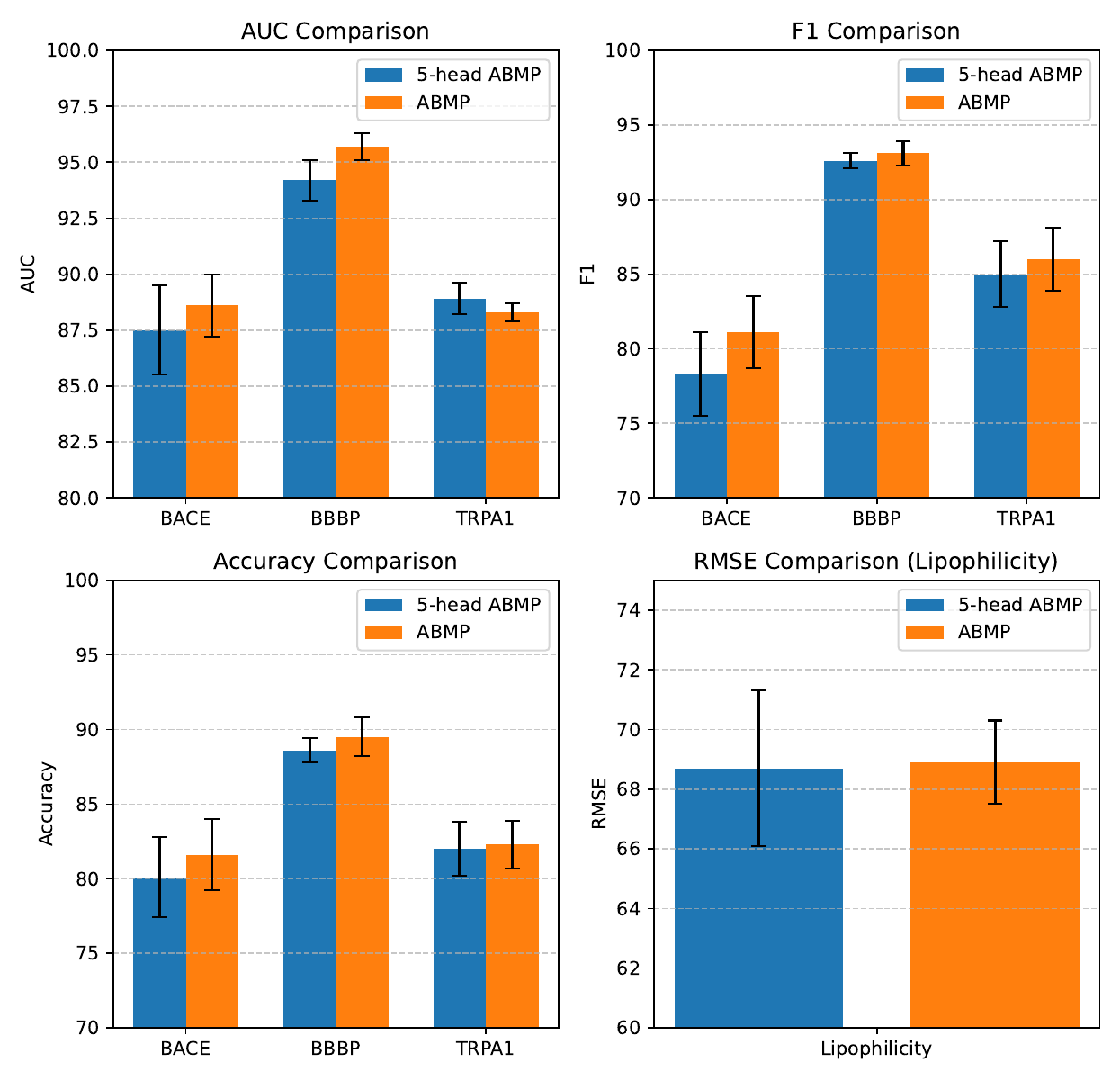}
  \caption{Comparison of ABMP and 5-head ABMP across three classification datasets (BACE, BBBP, TRPA1) and one regression dataset (Lipophilicity). Across all classification datasets (BACE, BBBP, TRPA1), 5-head ABMP and ABMP exhibit comparable performance. The subplots display performance metrics: Area Under the Curve (AUC comparison), F1 score values, Accuracy for classification tasks, and (D) Root Mean Square Error (RMSE) for regression. Bars represent the mean values, and error bars indicate the margin of error (MOE) with a confidence of 95\%.
}
  \label{fgr:9}
\end{figure}

To finalize the ABMP model analysis, we tested if adding a multi-head attention protocol would enhance the ABMP model further (Fig. ~\ref{fgr:9}), but no significant improvement was obtained using five heads and in fact, in the BBBP case showed statistically higher AUC value than the 5-head ABMP model. This finding reinforces our earlier observation regarding that simpler models could be more beneficial than complex ones for molecular property predictions. In small molecular graphs with limited structural variation, adding multiple heads could mean learning on redundant patterns and risking overfitting, especially if the dataset is not diverse or large.

Comparing the two top performing models,\textbf{ ABMP vs. BMP + SN}, the best model is ABMP for all metric scores in the BACE case, while BMP + SN had a higher AUC value for the TRPA1 dataset. We tried to adjust the classification threshold using the \textit{Youden index} (true positive rate -false positive rate) for both models, but we obtained similar F1 and accuracy scores, which discarded the need for a threshold adjustment in these specific cases. 

\begin{figure*}[H]
\centering
  \includegraphics[height=12 cm]{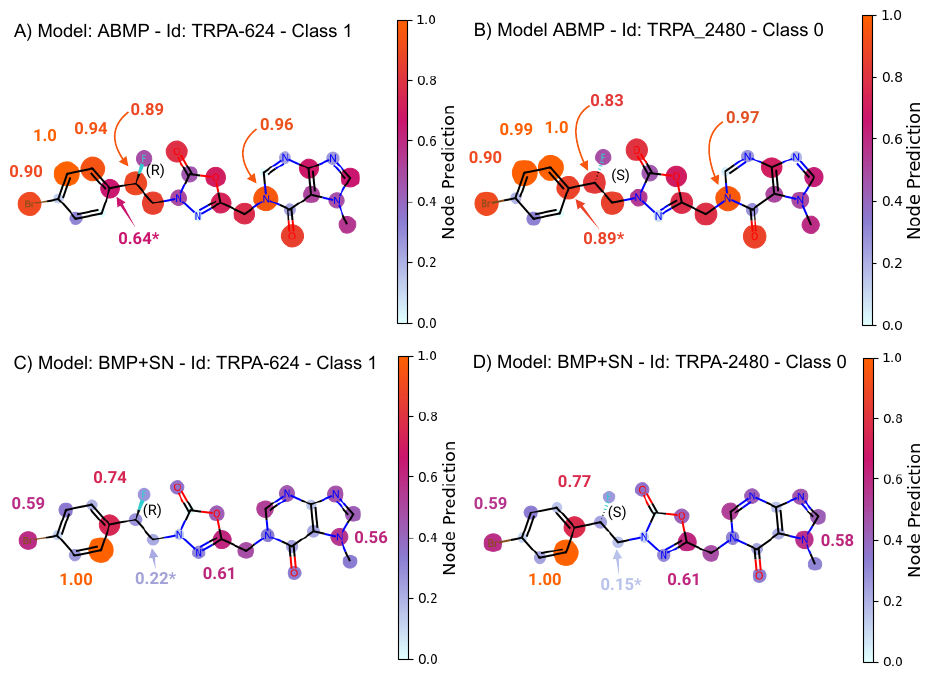}
  \caption{Colormaps for the normalized node-level outputs of two top performing models, ABMP and BMP + SN, corresponding to the challenging task of discriminating activity dictated by a chiral difference. The numerical highlighted values correspond to the top 5 highest min-max normalized node-level outputs and those who have remarked difference between stereoisomers which are highlighted with an asterisk (*). A) Molecular colormap of the TRPA-624 molecule derived from the ABMP model. This molecule is a class ’1’ and is characterized by the presence of an F(R) stereocenter. B) ABMP prediction for TRPA-2480, a stereoisomer of TRPA-624, where F(s) makes this compound inactive. C) BMP+SN node-level output for TRPA-624. D) BMP+SN node-level output for TRPA-2480. 
}
  \label{fgr:10}
\end{figure*}

Since the main distinction between the two models lies in the node block, we analyzed the molecular colormaps, which project node-level outputs onto corresponding atoms in each molecule and compare these two top-performing architectures. We focused on challenging cases of the TRPA1 dataset in a family of oxidiazole purines, for example where activity is dictated by a single chirality difference (BD-624 \& BD-2480)(Fig.~\ref{fgr:10})). 

Both ABMP and BMP + SN have a stable colormap patterns for the pair of molecules, which is a good indication of consistent learned embeddings at the node-level. In these cases, the ABMP was able to identify the relevant moiety for activity prediction in the samples, with higher node relevance assigned to the pertinent chiral carbon that dictates activity.

However, the difference between node embeddings is probably more significant for distinguishing classes. Ideally, in the (S)-enantiomer, the chiral carbon should receive lower node relevance, with nearby atoms also affected due to the spatial rearrangement caused by this subtle change in chirality. In fact, both models showed that the largest differences in raw node-level output occurred at the chiral carbon and its adjacent connected carbons, consistent with this structural perturbation.

To further investigate differences in node-level predictions, we aligned node outputs for pairs of molecules with the same number of atoms that differed by class and subtle variations in atomic position or element type. We then summed the node outputs for each molecule and calculated the differences between paired molecules to assess the ability of the two top performing models to distinguish them. This analysis was applied to four representative cases within the oxadiazole–purines family. The resulting pairwise node-output differences are summarized in Table~\ref{tbl:pairwise_differences}. Across all four cases, ABMP shows larger node-output differences compared to BMP+SN, suggesting that it is more sensitive to subtle structural variations such as chirality or positional isomerism.
\begin{table}[H]
\centering
\small
\caption{Comparison of node output differences between molecule pairs with equal atom counts and associated structural variations for the BMP+SN and ABMP models working with the TRPA1 dataset.}
\label{tbl:pairwise_differences}
\renewcommand{\arraystretch}{1.3} 
\begin{tabular}{llll}
    \hline
    \textbf{Pair of Molecules} & \textbf{BMP+SN} & \textbf{ABMP} & \textbf{Structural Difference} \\
    \hline
    BD-2480/BD-624   & $10\times 10^{-3}$  & $30\times 10^{-3}$  & Single chiral carbon \\
    BD-2415/BD-682   & $90\times 10^{-3}$  & $240\times 10^{-3}$ & Halogen \& position \\
    BD-1279/BD-2038  & $30\times 10^{-3}$  & $70\times 10^{-3}$  & Halogen position \\
    BD-1280/BD-1903  & $110\times 10^{-3}$ & $280\times 10^{-3}$ & Methyl position \\
    \hline
\end{tabular}
\end{table}

GATs improve MPNNs by learning to differentially weight neighboring nodes during message passing, allowing the model to focus on the most informative connections rather than treating all neighbors equally. This improves the ability to capture important structural and feature-level differences, manifested as increased differences in our node-level outputs ~\cite{khalil-GAT}. Also, learning can be stabilized by decoupling attention computation from message embeddings, reducing dependency on dynamically evolving latent features, while retaining interpretability through weights grounded in the raw node features.

Considering node-level outputs and global predictions together, we can conclude that the ABMP model is the most accurate, with a robust capacity to identify the vast majority of positive classes. The colormap results highlight this model for future applications where node-level prediction could aid at lead-optimization phases, identifying hot-spots for chemical transformations.

\subsection{Performance Comparison}

We contrasted our findings with 12 other studies falling into three categories related to different aspects of our models and offer a complete view of the state-of-the-art in molecular predictions:

\begin{itemize}
    \item \textbf{Baseline and/or High Scoring Methods}: Graph Attention Network (GAT)\cite{velickovic_graph_2018}, Graph Convolutional Networks (GCN)\cite{duvenaud_convolutional_2015}, MPNN\cite{gilmer_neural_2017}, Directed MPNN (D-MPNN)\cite{yang_analyzing_2019} and Dense Neural Networks working with fingerprints (DNN)\cite{deng_xgraphboost_2021}.
    
    \item \textbf{Advanced architectures}: Pharmacophoric-constrained Heterogeneous Graph Transformer (PharmHGT)\cite{jiang_pharmacophoric-constrained_2023}, Graph representation from self-supervised message passing transformer using the large version (GROVER)\cite{rong_self-supervised_2020} and Graph Structure Learning Molecular Property Prediction (GSL-MPP)\cite{zhao_molecular_2024}.
    
    \item \textbf{Attention mechanism with MPNN}: DumplingGNN\cite{xu_dumpling_2024} and Attentive-FP\cite{zhao_molecular_2024}. 
    
    \item \textbf{3D geometrical GNNs}: Uni-Mol\cite{zhou_uni-mol_2022}and 3DGCL\cite{moon_3d_2023}. 
\end{itemize}

\begin{figure*}[h]
\centering
  \includegraphics[height=11 cm]{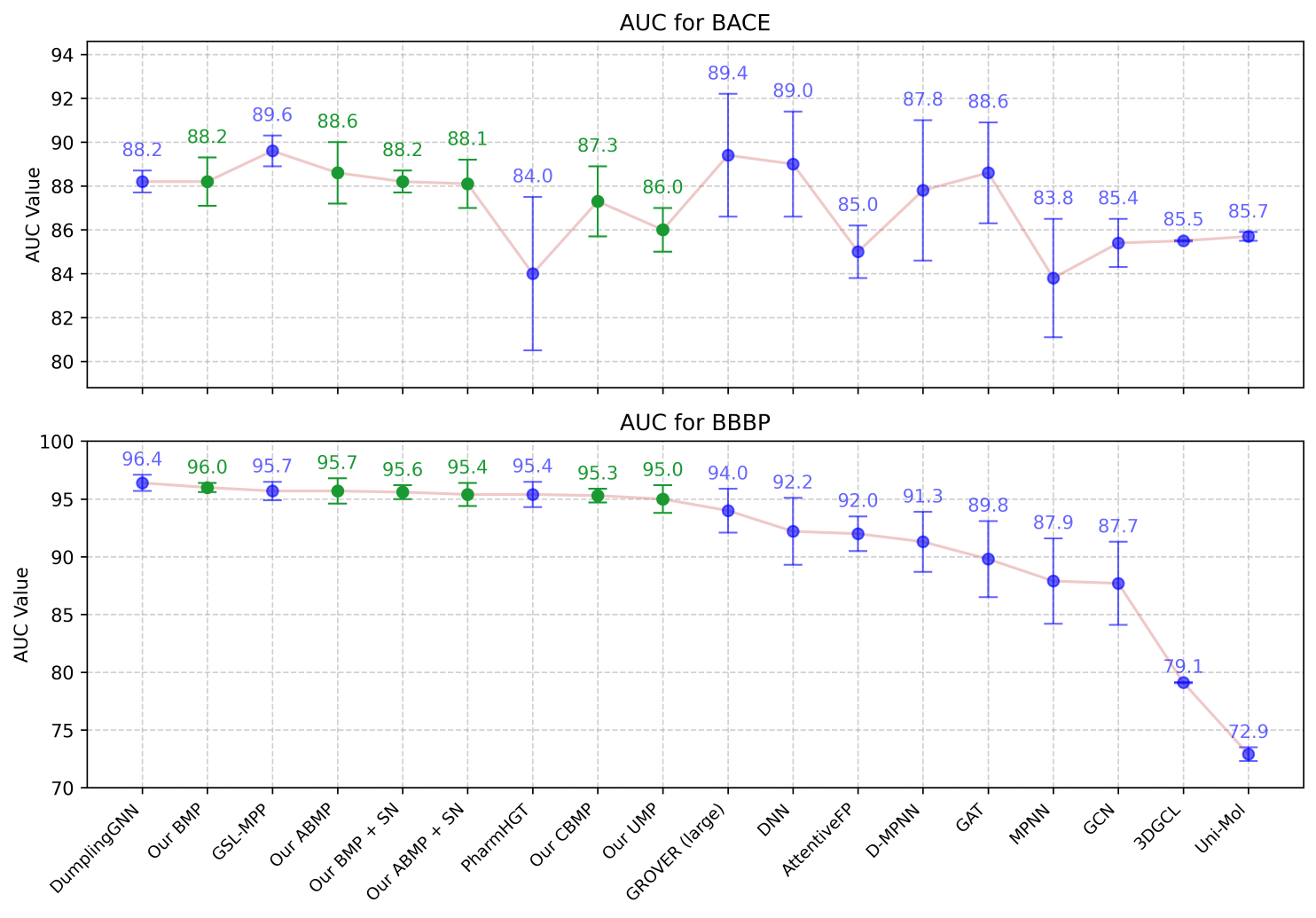}
  \caption{Comparative plot of Area Under the Curve (AUC) performance for different models on two datasets: BACE (top panel) and BBBP (bottom panel). The models were selected for their significance in the field, showcasing foundational architectures, incorporation of attention mechanisms, 3D embeddings, and novel architectural designs. The specific AUC values for each model were gathered from established sources: AttentiveFP\cite{zhao_molecular_2024}, D-MPNN\cite{yang_analyzing_2019}, DNN\cite{jiang_could_2021}, GAT\cite{jiang_could_2021}, GCN\cite{zang_hierarchical_2023}, GROVER (large)\cite{rong_self-supervised_2020}, GSL-MPP\cite{zhao_molecular_2024}, MPNN\cite{jiang_could_2021}, Uni-Mol\cite{zhou_uni-mol_2022}, 3DGCL\cite{moon_3d_2023}, DumplingGNN\cite{xu_dumpling_2024} and PharmHGT\cite{jiang_pharmacophoric-constrained_2023}. 
}
  \label{fgr:12}
\end{figure*}

The (Fig.~\ref{fgr:12}) summarizes the AUC values retrieved from different sources and is compared with the values we obtained from all our models using the two MoleculeNet benchmark datasets. This comparison demonstrated that our top models have strong AUC values with reduced error margins compared to other models reported elsewhere.

\begin{figure*}[h]
\centering
  \includegraphics[height=6.5 cm]{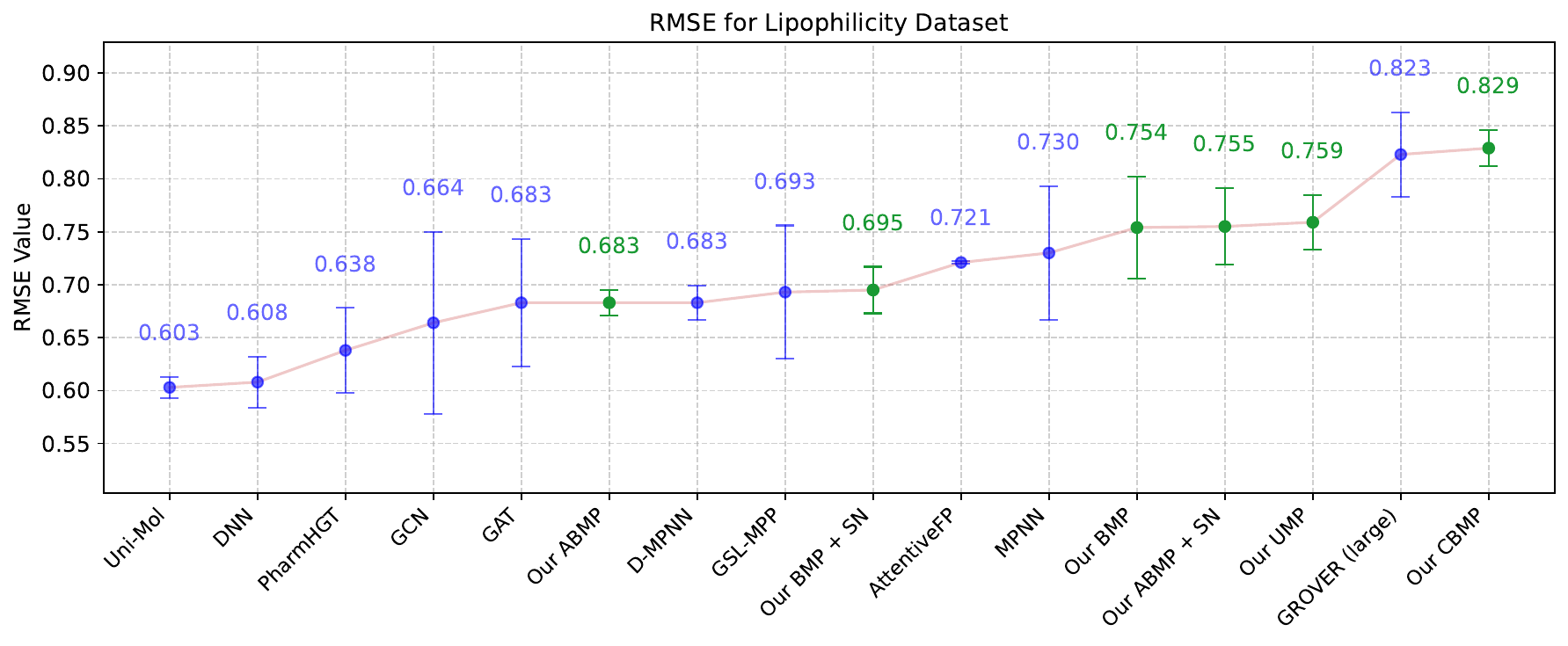}
  \caption{Comparative plot of RMSE performance for different models on the Lipophilicity (MoleculeNet) dataset. The models were selected for their significance in the field, showcasing foundational architectures, incorporation of attention mechanisms, 3D embeddings, and novel architectural designs. The specific RMSE values for each model were gathered from established sources: AttentiveFP\cite{jiang_dgcl_2024}, D-MPNN\cite{jiang_dgcl_2024}, DNN\cite{jiang_could_2021}, GAT\cite{jiang_dgcl_2024}, GCN\cite{jiang_dgcl_2024}, GROVER (large)\cite{jiang_could_2021}, GSL-MPP\cite{zhao_molecular_2024}, MPNN\cite{zhao_molecular_2024}, Uni-Mol\cite{zhou_uni-mol_2022}, DGCL\cite{jiang_dgcl_2024} and PharmHGT\cite{jiang_pharmacophoric-constrained_2023}. 
}
  \label{fgr:13}
\end{figure*}

We then evaluated our models on the Lipophilicity dataset and compared their performance against previously reported models (Fig.~\ref{fgr:13}).

Our results show that ABMP achieved performance comparable to GAT in the case of BACE and Lipophilicity, and outperformed it in the BBBP dataset. Additionally, both ABMP and BMP+SN surpassed classical MPNN across all datasets, and all our models achieved higher results for the classification tasks in this same reference. For regression, ABMP achieved the best score among our models, with an RMSE of $0.683 \pm 0.016$, comparable to D-MPNN, which similarly incorporates edge features into message processing but has a different strategy for the passing. However, in the classification tasks, most of our models surpassed the D-MPNN. These findings demonstrate that our bidirectional message-passing combined with edge-aware attention outperforms their precursor architectures.

ABMP (BBBP) and BMP+SN (Lipophilicity) obtained similar performance to GSL-MPP, a complex architecture. GSL-MPP uses a dual-graph approach: one graph for molecular structures and another where molecules act as nodes connected via similarity matrices, effectively doubling the computational complexity to match our single-graph models~\cite{zhao_molecular_2024}. Similarly, our models outperformed GROVER, a pre-trained transformer-based MPNN trained on over 10 million compounds, particularly on BBBP~\cite{rong_self-supervised_2020}. This aligns with prior findings that pre-trained transformer-based models can be outperformed by MPNN-based architectures in molecular property predictions~\cite{abbassi_gmpp-nn_2024}. This reinforces that added architectural complexity, even the large-scale pre-trained or 3D-intensive models, does not necessarily yield better performance. This same observation can be made comparing GAT and even descriptor-based models like DNN outperforming more complex approaches such as Uni-Mol and 3DGCL in classification tasks. 

Among related models, DumplingGNN most closely resembles the ABMP. Both incorporate attention mechanisms and 3D embeddings, and both achieved similar classification performance. However, DumplingGNN employs sequential MPNN layers followed by GAT with SAGEconv, while ABMP integrates message passing directly with attention and uses simpler global max-pooling. DumplingGNN lacks of 3D inputs for BACE, likely explaining its lower performance, underscoring the importance of 3D features, as supported by our feature selection and the superior performance of UMP (with 3D features) over 2D-featurized MPNN.

This results also support the notion that 3D coordinates alone are insufficient to guarantee superior performance, as demonstrated by Uni-Mol and 3DGCL underperforming relative to DNN. These models leverage large-scale pretraining and explicit geometric encoding but fail to outperform simpler alternatives working with fingerprints. This suggests that, for molecular property prediction, representing chemically relevant features (e.g., buried volume or electronic effects) is often more effective than full 3D coordinate modeling, particularly for global property prediction tasks.

\section{Conclusion}

In this study, we adopted a minimalist approach to message-passing by deliberately excluding self-node representations and incrementally adding key components from the three main GNN paradigms. This step-by-step framework enabled us to isolate and evaluate the contribution of each architectural element to model performance, ensuring that only essential features were retained in the final designs.

Our results indicate that the internal dataset structural dissimilarity dictates the need for architectural complexity. For instance, the BBBP dataset, which exhibits low structural dissimilarity, achieved state-of-the-art performance using our simplest model (BMP). This highlights that simpler architectures can outperform more complex ones under the right conditions, contrasting with the prevailing trend in GNNs that often favors increased architectural depth and sophistication.

A central focus of this work was feature selection. We found that element-like features (e.g., atomic number) often hurt performance by overemphasizing certain atoms (e.g., carbon), which are prevalent in organic molecules. Instead, we prioritized tailored spatial features over traditional 2D descriptors, which consistently improved model performance across all tasks. Notably, we showed that 3D-inspired features computed from 2D molecular graphs can match the predictive power of features derived from fully optimized 3D conformations, offering a simpler and faster alternative for molecular modeling.

We also explored two strategies for handling directionality in message-passing. Our bidirectional model (BMPs) outperformed the undirected variant (UMP), both in performance and parameter efficiency. We attribute this to UMP’s reliance on mean-pooling, which tends to dilute informative signals when averaging across duplicated edges.

Across all models, we observed higher F1 scores than accuracy on the TRPA1 and BBBP datasets, reflecting class imbalance that favors the active class. Since F1 prioritizes the quality of positive class predictions, it provided a more realistic assessment of model effectiveness in these cases. Additionally, we found that the the number of features to descrribe a dataset is related to the structural diversity, where a highly diverse dataset would require more descriptors to allow efficient learning during training.

Finally, we evaluated the role of convolution and attention mechanisms. The CBMP model showed no improvement over simpler variants, likely because degree penalization was unnecessary due to the low connectivity and exclusion of hydrogens in our molecular graphs, which resulted in a normal-like degree distribution. In contrast, the ABMP model showed clear advantages in both classification and regression tasks. Its attention mechanism improved node-level focus on functionally relevant atom groups, as confirmed by UMAP visualizations and molecular node-outputs projections onto molecular heatmaps. These results suggest that ABMP is the most promising architecture for drug discovery applications among all models tested.

\section{Limitations and Future Work}

Although the proposed framework achieves state-of-the-art results with reduced complexity, there are inherent limitations and scalability challenges that must be addressed to extend its applicability to larger and more diverse datasets.

The cardinality problem in GATs arises because softmax-based attention normalizes neighbor contributions through exponential competition, followed by scatter-sum aggregation that compounds the effect~\cite{zhang_improving_2020}. For high-degree nodes, this dual operation diminishes attention weights and weakens aggregated signals, while also biasing the shared attention vector toward patterns prevalent in densely connected nodes. This ultimately underrepresents important neighbors in high-degree nodes and overemphasizes low-degree nodes in certain cases. In this work, we mitigated this issue by employing scatter-max aggregation, which removes the normalization step. However, this approach limits applicability for tasks requiring multiple aggregated contributions, such as HOMO–LUMO gap prediction. As part of future work, we plan to address this limitation by using datasets tailored to quantum mechanical properties (e.g., HOMO–LUMO gap) and applying convolution-based normalization directly on attention weights, thereby incorporating degree information without resorting to relative softmax normalization.

This study employs a single message-passing iteration across all models, leaving the effect of multiple passes unexplored. Treating the number of passes as a hyperparameter could improve performance for tasks that require deeper information propagation. Similarly, we used a fixed training schedule of 50 epochs, which may not be optimal. Future work could benefit from tuning the number of epochs or incorporating early stopping criteria to avoid overfitting and improve convergence efficiency.

Multi-headed attention did not outperform single-head attention in this study, though only five heads were tested. Further experiments are needed to evaluate whether the number of attention heads significantly influences performance.

While our results show that 2D molecular graphs supplemented with select 3D descriptors can maintain predictive accuracy, our study relied on force-field–optimized conformations, which are approximations and may not fully capture true experimental or quantum-optimized 3D geometries. Consequently, we did not rigorously assess performance in scenarios where highly reliable 3D structures are available. Future work will involve testing on datasets with experimentally determined or high-level QM-derived conformations to better evaluate the benefits of accurate 3D geometries.

The dual-goal hyperparameter tuning strategy required significant trials (e.g., 200 for TRPA1) and struggled with convergence on low-diversity datasets such as BBBP. Exploring alternative optimization objectives or simplifying the approach to a single-direction strategy warrants further investigation.

Inspired by the success of ABMP and BMP + SN, the continuation of this research project will involve building an adaptive architecture that composes the right message for a given dataset. Hence, the model should be capable of internally selecting the inclusion of self-nodes, edge/node/global embeddings and attention mechanisms as part of the training process. On a different aspect, while 3D features such as buried volume and radius of gyration improved model performance, further exploration into more advanced 3D descriptors or spatial embeddings is warranted. 

The ability of ABMP to highlight node-level relevance suggests their potential for real-world applications in lead optimization and drug design. In this work, we focus exclusively on the message and use a single pass, leaving the optimization of the number of passes unexplored. A potential continuation of this work could involve treating the number of passes as a hyperparameter to be optimized. Moreover, we are intrigued to explore why the use of multi-headed attention in ABMP did not yield better results than a single-head. A prospective study could include whether the number of heads is a decisive factor in performance. 

\section*{Author contributions}

Research work and writing made by ACCL. Review made by REA.

\section*{Data availability}

The code developed for this study, including all model training scripts, dataset processing routines, and evaluation examples for molecular property prediction using our BMP variations are available at \href{https://github.com/chemdesign-accl/BMPs}{https://github.com/chemdesign-accl/BMPs}. The version of the code employed for this study is \texttt{v1.0}. 

To reproduce the results we obtained for each model and dataset for our final predictive power results (Fig.~\ref{fgr:5} and Fig.~\ref{fgr:7}) run the python script \texttt{BMPNNs/examples/evaluate\_internal\_split.py} using the listed hyperparameters for each model and dataset combination available in Supplemental Information, Appendix D.

All datasets analyzed in this work are derived from publicly available sources, including BACE and BBBP from MoleculeNet and TRPA1 from BindingDB.

\section*{Acknowledgements}

Alma C. Castañeda-Leautaud gratefully acknowledges financial support for this research by the Fulbright U.S. Student Program, which is sponsored by the U.S. Department of State and COMEXUS. Its contents are solely the responsibility of the author and do not necessarily represent the official views of the Fulbright Program, the Government of the United States, or COMEXUS.

We extend our gratitude to the Midwest AviDD Center and its research team, including Thomas Bannister, Ph.D., Donghoon Chung, Ph.D., and Ambuj Srivastava, Ph.D., for the valuable discussion that contributed to the advancement of this project.

Portions of the code were developed with the assistance of large language models (ChatGPT, OpenAI), and subsequently verified and adapted by the authors.

\bibliographystyle{unsrtnat}   
\bibliography{main}

\end{document}